\RequirePackage{fix-cm}
\documentclass[twocolumn]{svjour3}          
\smartqed  

\usepackage{cite}
\usepackage[misc]{ifsym} 
\usepackage{algorithm}
\usepackage{algpseudocode}
\usepackage{amsmath}
\usepackage{amssymb}
\usepackage{graphicx}
\usepackage{subcaption}
\usepackage{footnote}

\usepackage[dont-mess-around]{fnpct}

\usepackage{scalerel}
\usepackage{tikz}
\usetikzlibrary{svg.path}
\definecolor{orcidlogocol}{HTML}{A6CE39}
\tikzset{
  orcidlogo/.pic={
    \fill[orcidlogocol] svg{M256,128c0,70.7-57.3,128-128,128C57.3,256,0,198.7,0,128C0,57.3,57.3,0,128,0C198.7,0,256,57.3,256,128z};
    \fill[white] svg{M86.3,186.2H70.9V79.1h15.4v48.4V186.2z}
                 svg{M108.9,79.1h41.6c39.6,0,57,28.3,57,53.6c0,27.5-21.5,53.6-56.8,53.6h-41.8V79.1z M124.3,172.4h24.5c34.9,0,42.9-26.5,42.9-39.7c0-21.5-13.7-39.7-43.7-39.7h-23.7V172.4z}
                 svg{M88.7,56.8c0,5.5-4.5,10.1-10.1,10.1c-5.6,0-10.1-4.6-10.1-10.1c0-5.6,4.5-10.1,10.1-10.1C84.2,46.7,88.7,51.3,88.7,56.8z};
  }
}
\newcommand\orcidicon[1]{\href{https://orcid.org/#1}{\mbox{\scalerel*{
\begin{tikzpicture}[yscale=-1,transform shape]
\pic{orcidlogo};
\end{tikzpicture}
}{|}}}}
\usepackage{hyperref} 

\usepackage{etoolbox}
\newcommand*{\affaddr}[1]{#1} 
\newcommand*{\affmark}[1][*]{\textsuperscript{#1}}

\usepackage{fancyhdr}
\begin{document}
\sloppy

\title{Knowledge-Based Hierarchical POMDPs for Task Planning}

\author{Sergio A. Serrano\affmark[1] \and
        Elizabeth Santiago\affmark[2] \and 
        Jose Martinez-Carranza\affmark[1,3] \and
        Eduardo Morales\affmark[1,4] \and
        L. Enrique Sucar\affmark[1]
        }
\authorrunning{S. A. Serrano et al.} 

\institute{(\Letter) Sergio A. Serrano \orcidicon{0000-0002-4994-9052} \\
        sserrano@inaoep.mx\\ \\
        Elizabeth Santiago \orcidicon{0000-0001-5801-4796} \\
        eliza.stgo@gmail.com\\ \\
        Jose Martinez-Carranza \orcidicon{0000-0002-8914-1904} \\
        carranza@inaoep.mx\\ \\
        Eduardo Morales \orcidicon{0000-0002-7618-8762} \\
        emorales@inaoep.mx\\ \\
        L. Enrique Sucar \orcidicon{0000-0002-3685-5567} \\
        esucar@inaoep.mx \\ \\
        \affaddr{\affmark[1] Instituto Nacional de Astrof\'isica, \'Optica y Electr\'onica, Computer Science Department, Sta. Ma. Tonantzintla, Puebla, M\'exico}\\
        \affaddr{\affmark[2] Instituto Nacional de Enfermedades Respiratorias, Laboratory of Computational Biology, Ciudad de M\'exico, M\'exico}\\
        \affaddr{\affmark[3] University of Bristol, Computer Science Department, Bristol, UK}\\
        \affaddr{\affmark[4] Centro de Investigaciones en Matem\'aticas, Guanajuato, Guanajuato, M\'exico}
        }

\date{Received: date / Accepted: date}

\maketitle
\begin{abstract}

The main goal in task planning is to build a sequence of actions that takes an agent from an initial state to a goal state. In robotics, this is particularly difficult because actions usually have several possible results, and sensors are prone to produce measurements with error. Partially observable Markov decision processes (POMDPs) are commonly employed, thanks to their capacity to model the uncertainty of actions that modify and monitor the state of a system. However, since solving a POMDP is computationally expensive, their usage becomes prohibitive for most robotic applications. In this paper, we propose a task planning architecture for service robotics. In the context of service robot design, we present a scheme to encode knowledge about the robot and its environment, that promotes the modularity and reuse of information. Also, we introduce a new recursive definition of a POMDP that enables our architecture to autonomously build a hierarchy of POMDPs, so that it can be used to generate and execute plans that solve the task at hand. Experimental results show that, in comparison to baseline methods, by following a recursive hierarchical approach the architecture is able to significantly reduce the planning time, while maintaining (or even improving) the robustness under several scenarios that vary in uncertainty and size.


\end{abstract}

\section*{Category (1).}

\section{Introduction} \label{sec:intro}
In recent decades, robots have become less a character only found in fictional stories, and more a realistic solution to the constantly growing demand of products and services, that industries today are challenged to suffice. This can be observed in the progression of their role in the private sector, going all the way from automated production lines \cite{eversheim1982recent,chin1982automated}, to warehouse robots \cite{poudel2013coordinating} and, more recently, as service robots designed to collaborate in duties related to health care, education and business \cite{pandey2018mass}. A key component in autonomy, which the latter group of robots requires, lies in the capacity of a system to solve a diversity of tasks by itself.

\paragraph{}
In order to solve tasks autonomously, robots must be equipped with the knowledge (or the capacity to acquire it) about the structure of the class of problems they are being designed to solve, as well about how their actions can change the state of world. The degree of faithfulness/consistency of the representation, and the actual problem, has a great impact in the success of the plans a robot generates to solve the task at hand. For instance, in the real world, for most of the actions a robot may perform, there is a chance of obtaining an unexpected result. Similarly, sensors are always prone to produce incorrect measurements, in some degree. Thus, in the context of planning robots, partially observable Markov decision processes (POMDP) \cite{puterman2004markov,kaelbling1998planning} have been widely used to address task planning problems, since they are able to model the stochastic dynamics of acting and measuring processes, such as the actuators and sensors in a service robot.

\paragraph{}
However, considering that solving a POMDP is a PSPACE-\textit{hard} problem \cite{papadimitriou1987complexity}, and that it is common for robotic platforms to have a significant amount of degrees of freedom (\textit{e.g.}, the position, speed and torque of each articulation), solving a POMDP that models the state variables of a whole robot is very expensive (in terms of computational time), and becomes unfeasible for most applications. In order to mitigate the computational cost, different strategies have been explored from which the hierarchical approach, in tandem with Markov decision processes (MDP) \cite{dietterich2000hierarchical,hengst2002discovering}, is one the most employed techniques because of its capacity to reduce the overall complexity of a problem, by decomposing it into a hierarchy of smaller problems.

\paragraph{}
Hence, in this paper we propose a task planning architecture that combines a hierarchical approach and POMDPs, with a knowledge-based scheme, to solve decision-making problems. The architecture defines a knowledge base structure that a designer can use to encode descriptions of the skill sets of a robot, as well of the structure of its environment, in a way that promotes the modularity and reuse of information. Then, the architecture initializes by building a hierarchy of POMDPs, based on the descriptions in the knowledge base and a recursive definition of a POMDP. Afterwards, the architecture is ready to accept task requests from a user, which it solves by generating (and executing) multi-resolution plans that are built upon the hierarchy of POMDPs (see Fig. \ref{fig:gen-workflow}).

\paragraph{}
By modeling plans at several levels of granularity, the architecture facilitates the interpretation of the decisions the agent makes. Moreover, since abstract actions are modeled as POMDPs defined over a subregion of the state space, during the execution of plans, the architecture employs an entropy-based statistic so that the agent is better informed about its current state. To evaluate the proposed architecture, we used a mobile robot navigation domain. Experimental results show that, by employing a recursive hierarchical approach, the system is capable of outperforming the baseline methods, in terms of computing time and effectiveness to reach goal locations. After submitting the architecture to several scenarios (that differ in size and overall uncertainty), it consistently exhibited a robust behavior.

\paragraph{}
The rest of the paper is structured as follows, Section \ref{sec:rw} summarizes the main advances in planning systems that combine POMDPs with a hierarchical approach, Section \ref{sec:gen-overview} presents the overall functionality of the proposed architecture and its three main stages. Section \ref{sec:kbc} describes the components of information that are required to specify the knowledge base (first stage), then, in Section \ref{sec:ai} is presented the procedure that initializes the architecture (second stage), followed by Section \ref{sec:ao} which describes the operation of the architecture (third stage), \textit{i.e.}, when the agent is ready to solve tasks. Section \ref{sec:exp} shows the experiments and results obtained from the evaluation of the proposed architecture, and finally in Section \ref{sec:conc-fw} the conclusions and ideas for future work are presented.

\section{Related work}\label{sec:rw}
To endow a machine with the capacity to make decisions on its own has probably been one of the most exciting problems in artificial intelligence for many years. Hence, in an effort to overcome this challenge, a variety of approaches have been proposed. For instance, \cite{fikes1971strips} introduced STRIPS, a problem solving system based on first-order logic, which was widely employed to solve theorem-proving and planning problems. Eventually, systems with a greater expressive power were proposed, such as PDDL \cite{mcdermott1998pddl,fox2003pddl2}, which enabled designers to model planning problems in terms that were more concise and human-like. However, these approaches lacked the capacity to model problems in which events have several possible outcomes, such as robots interacting with people and with physical objects. Thus, researchers have looked for frameworks that are capable of modeling stochastic events, \textit{e.g.}, PPDDL \cite{younes2004ppddl1} which is the probabilistic extension to PDDL, and MDP \cite{puterman2004markov} which are one of the most used models to solve sequential decision making problems.

\paragraph{}
Moreover, since sensors are susceptible to produce wrong measurements (as any measuring system), robots cannot blindly trust their sensing devices to evaluate the current state of the environment. Thus, frameworks that assume to know (without error) the state of the system, like PPDDL or MDP, fall short to suffice the requirements of a planning robot. Nonetheless, there is a generalization of the MDP, known as partially observable MDP (POMDP) \cite{puterman2004markov,kaelbling1998planning}, that relaxes this assumption and models the measurement error a robot might have by means of a set of probability distributions. However, one of the major drawbacks POMDPs have is the high computational cost required to compute an optimal policy for them \cite{papadimitriou1987complexity}.

\paragraph{}
Despite the significant amount of research that has been done in the area of approximate solving algorithms \cite{pineau2003point,spaan2005perseus,kurniawati2008sarsop}, solving POMDPs remains intractable for most real-world task planning problems. Hence, alternative ways to mitigate the burden of computing near-optimal policies in less time have been explored, among which, approaches that decompose the original problem into a hierarchy of smaller ones have shown to be a promising solution. For example, \cite{pineau2001hierarchical,pineau2002integrated} proposed a hierarchical POMDP scheme in which the hierarchical structure is provided by a human designer in the form of a hierarchy of actions, \textit{i.e.}, a directed acyclic graph in which a set of children nodes represent all the actions that might be required to perform the parent action. By doing so, each POMDP in the hierarchy of actions had a reduced action space (with respect to the original one), which reduced in several orders of complexity the time required to compute a solution, in comparison to a POMDP with no decomposition.

\paragraph{}
On the other hand, \cite{theocharous2001learning,theocharous2002hierarchical} introduced hierarchies of POMDPs as an extension of hierarchical hidden Markov models (HHMM) \cite{fine1998hierarchical}, in which the hierarchical decomposition of the problem is provided in the form of a hierarchy of states (state abstraction). In contrast to \cite{pineau2001hierarchical}, it is the state space the one being reduced in each local POMDP in the hierarchy, and their system employed sequences of observations to train the overall model. In this way, after evaluating in a hall navigation domain, their system planned considerably faster than a regular POMDP, showing how feasible it can be to exploit the hierarchical structure of the environment (in this case in the form of interconnected halls) to learn a policy from a sequence of observations.

\paragraph{}
Recently, another form of hierarchical planning involving POMDPs has been explored \cite{zhang2012asp+,hanheide2017robot,zhang2017dynamically,sridharan2018reba,zhang2020icorpp}, in which a multi-resolution model is constituted by two main levels of representation: a high level at which commonsense reasoning is usually performed, and a low level at which portions of the environment are modeled as POMDPs. In this way, the system generates high-level plans whose actions are converted and executed as POMDPs. In these works, similar to \cite{theocharous2001learning,pineau2001hierarchical}, the hierarchical structure of the environment is provided by a designer, which they use to establish an association between elements of the two levels of representation. By doing so, the system exploits high-level knowledge (\textit{e.g.}, books are usually found in the library, not in the kitchen) to generate well informed plans and leave the details, that are relevant for the execution of actions, to the POMDPs at the low-level representation.

\paragraph{}
In this paper, we propose a task planning architecture for service robotics. We use SPARC \cite{balai2013sparc} as a representation language to define the structure of a knowledge base that enables a designer to encode the capabilities of the robot, as well as hierarchical information of the environment, to build hierarchies of POMDPs. In contrast to \cite{pineau2001hierarchical,pineau2002integrated}, we employ state abstraction as a source of hierarchical information, which enables our system to build a hierarchy of abstract actions (as POMDPs) in an automatic bottom-up manner. In this way, abstract actions are inferred based on the low level description of the environment and the state abstraction (both provided by the designer). Moreover, unlike \cite{theocharous2001learning,theocharous2002hierarchical}, our architecture does not depend on a sequence of observations to train the overall hierarchy of POMDPs. This is particularly important for service robots, because of the diversity of tasks they are expected to encounter \cite{ingrand2017deliberation}, it becomes unrealistic to generate enough sequences of observations, given all the things a robot might be capable of doing (\textit{e.g.}, navigating, manipulating objects, speaking, etc.).

\paragraph{}
On the other hand, we take the idea of task planning architectures one step further by providing a separate description of the robot and the environment, from which a hierarchy of POMDPs is automatically built. In this way, if a designer has knowledge of the hierarchical composition of a problem, our architecture is capable of exploiting it, contrary to approaches like \cite{zhang2017dynamically,sridharan2018reba} that are restricted to two levels of resolution. Furthermore, as experimental results show, there are some tasks (such as navigation) in which a hierarchy of POMDPs offers a better solution than a two-level architecture, in terms of efficiency and effectiveness (see Section \ref{sec:exp}).

\section{General overview} \label{sec:gen-overview}
\begin{figure}[]
    \centering
    \includegraphics[width=\linewidth]{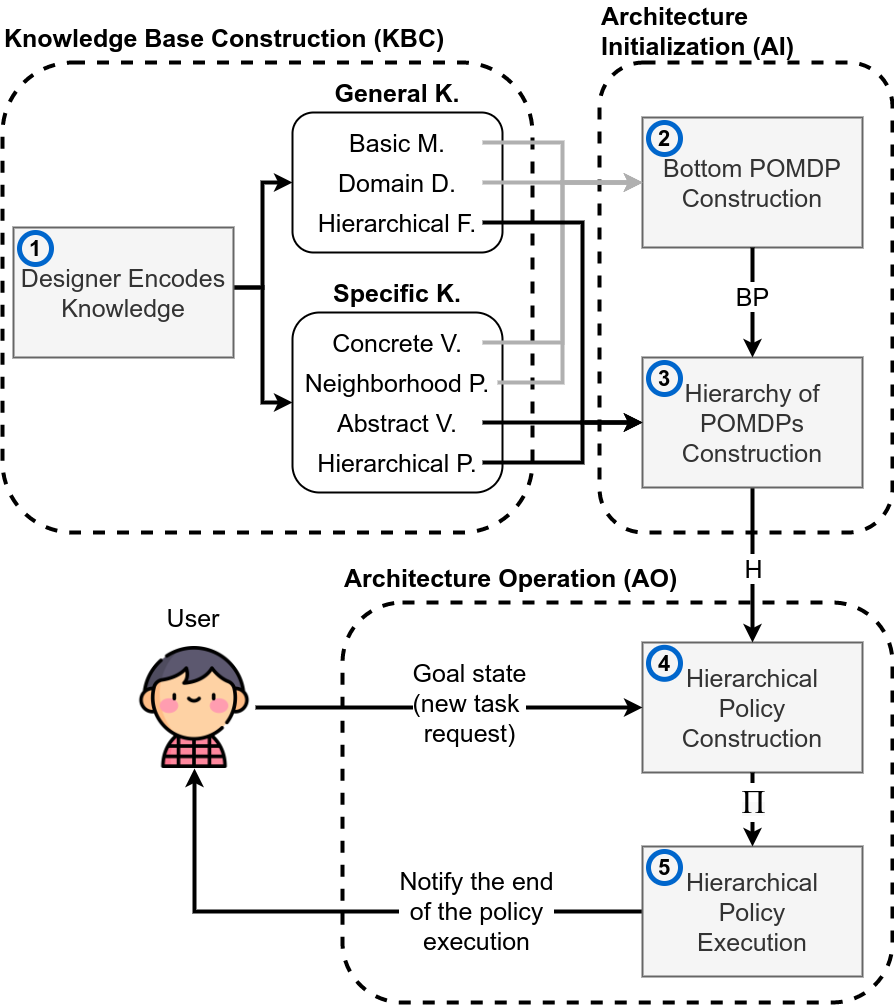}
    \caption{Overall workflow of the proposed architecture. In the KBC stage, a designer encodes a knowledge about the robot  and the environment. Next, in the AI stage, the basic modules, domain dynamics, concrete values and neighborhood pairs are employed to build a POMDP (BP) that models the interactions between the robot and the environment. Then, this POMDP is used, along with the hierarchical function, abstract values and hierarchical pairs to build the hierarchy of POMDPs (H). Finally, in the AO stage, the system is ready to receive task requests (in the form of goal states) from a human user. The task request and the hierarchy of POMDPs serve as input for the construction of a hierarchical policy ($\Pi$), which is executed to solve the requested task. Steps 4 and 5 can be executed as many times the user desires, however, one must wait for the robot to finish executing a hierarchical policy before requesting it to solve a new task.}
    \label{fig:gen-workflow}
\end{figure}

In order to solve task planning problems, the proposed architecture follows three stages: a) knowledge base construction (KBC), b) architecture initialization (AI), and c) architecture operation (AO). In the KBC stage, a human designer is required to encode, into the knowledge base (KB), domain specific information that describes the skill set of the robot and the particular scenario in which it will operate. Next, in the AI stage, the architecture builds a POMDP from the information in the KB, and uses a hierarchical description of the environment to build a hierarchy of POMDPs. Finally, in the AO stage, the architecture is ready to receive task requests, for which it builds and executes multi-resolution plans (based on the hierarchy of POMDPs). Both the KBC and AI stages are performed only once, while the AO stage is executed every time a task request is issued to the robot (see Fig. \ref{fig:gen-workflow}).

\section{Knowledge base construction} \label{sec:kbc}
The encoding of the KB, which is performed only once, consists in the designer providing general and specific knowledge that describes the environment that the robot will operate in, as well as the skill set it is equipped with. The general knowledge is specified by three components: basic modules, domain dynamics, and a hierarchical function. Whilst the specific knowledge is defined by four lists of elements: concrete values, abstract values, neighborhood pairs and hierarchical function pairs. Each of these components is presented in the following sections.

\begin{figure*}[t!]
    \centering
     \begin{subfigure}{0.26\textwidth}
         \centering
         \includegraphics[width=0.9\linewidth]{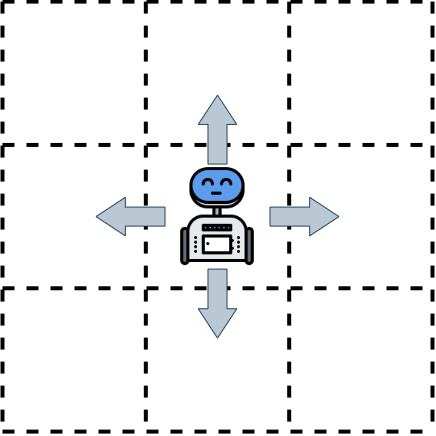}
         \caption{Set of actions}
         \label{fig:nav-a}
     \end{subfigure}
     \begin{subfigure}{0.26\textwidth}
         \centering
         \includegraphics[width=0.9\linewidth]{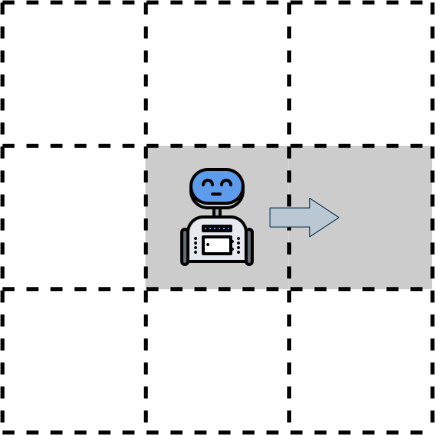}
         \caption{Transition distribution}
         \label{fig:nav-t}
     \end{subfigure}
     \begin{subfigure}{0.34\textwidth}
         \centering
         \includegraphics[width=0.9\linewidth]{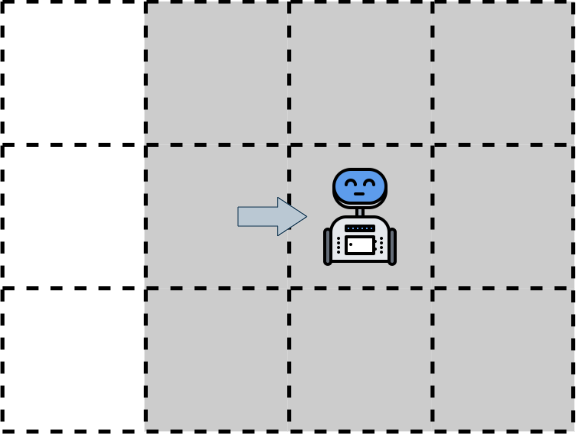}
         \caption{Observation distribution}
         \label{fig:nav-o}
     \end{subfigure}
        \caption{Example of a mobile robot navigation domain. The robot can move using any of its four actions: up, down, left and right. The transition distribution of every action consists of two cells: the current one and the target cell, as shown by the shaded cells in Fig. \ref{fig:nav-t}. Whereas, the observation distribution of every action consists of those cells covered by a $3 \times 3$ kernel that is centered in the cell that has been reached by the robot after executing an action, Fig. \ref{fig:nav-o} shows the cells covered by the kernel in gray.}
        \label{fig:nav-ato}
\end{figure*}

\paragraph{}
Furthermore, for illustration purposes, lets consider an example of a mobile robot that navigates within a closed discrete environment (see Fig. \ref{fig:nav-ato}). Despite this robot has a complete map of its environment, because it is an old platform and has been operating for many years, sometimes, its localization sensors fail and its wheels drift. Therefore, in order for the robot to plan, a POMDP comes as a suitable solution, as it is capable of modeling the uncertainty in the outcomes from its actions (see Figs. \ref{fig:nav-a} and \ref{fig:nav-t}), as well as the error in its measurements (see Fig. \ref{fig:nav-o}). This example will be referenced throughout the description of the proposed architecture.

\subsection{General knowledge}
In the general knowledge of the KB, for each skill set the robot has (\textit{e.g.}, navigation, object manipulation, etc.) a basic module must be specified. The domain dynamics must contain a description of how the actions the robot is capable of performing can interact with the environment, while the hierarchical function must specify with respect to which variable the state space can be abstracted into a hierarchy. Furthermore, since the general knowledge describes the skills of a robot, it can be reused in different environments, as long as the hardware of the robot remains the same.

\subsubsection{Basic module}
A basic module is defined by a tuple $\langle A_{BM},V_{BM},W_{BM},O_{BM},T_{BM},Z_{BM} \rangle$, where each element of the tuple is defined as follows.

\paragraph{}
$\boldsymbol{A_{BM}}$: Set of actions the basic module is able to perform (\textit{e.g.}, a basic module for a mobile navigation skill could have a set $A_{BM}=\{up,down,left,right\}$ to move between adjacent locations, see Fig. \ref{fig:nav-a}).

\paragraph{}
$\boldsymbol{V_{BM}}$: Set of state variables that actions in $A_{BM}$ are capable of modifying. Each action in $A_{BM}$ must be capable of modifying exactly one state variable, and each state variable must be modifiable by at least one action (\textit{e.g.}, $V_{BM}=\{robot\_loc\}$ in the mobile robot example, where the four actions modify the same variable, since $robot\_loc$ represents the location of the robot). By restricting actions to modify a single variable, the architecture promotes a modular design and reduces the amount of experiments required to estimate the transition distribution of an action, since only one variable needs to be monitored.

\begin{figure*}[t!]
    \centering
    \includegraphics[width=\textwidth]{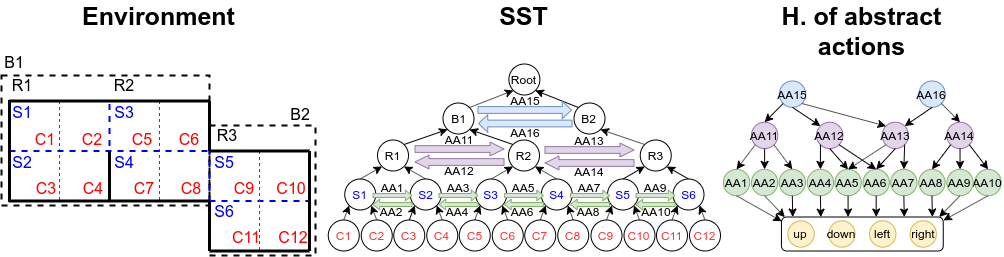}
    \caption{Example of an environment for the mobile robot presented in Fig. \ref{fig:nav-ato}. The environment is composed by 12 cells, which are abstracted into 6 sections, 3 rooms and 2 buildings to define the state space tree of the environment. Abstract actions (thick arrows) are built to transit between specific pairs of neighbor abstract states, and build the hierarchy of abstract actions (hierarchy of POMDPs). In this hierarchy, an abstract action can invoke any of its children actions.}
    \label{fig:example}
\end{figure*}

\paragraph{}
$\boldsymbol{W_{BM}}$: Set of sets of values for the state variables. For each $v_i$ in $V_{BM}$, there must be a set $w_i$ in $W_{BM}$ that contains all the values that $v_i$ can take. If the values for $v_i$ depend on the particular environment, then $w_i$ should be specified when the specific knowledge is provided (\textit{e.g.}, $w_1=\{cell_1,cell_2,...\}$ for $robot\_loc$, which depends on how many cells there are in the environment; in the case of the environment from Fig. \ref{fig:example} $w_1$ would be made of twelve cells).

\paragraph{}
$\boldsymbol{O_{BM}}$: Set of sets of observations for the state variables. For each $v_i$ in $V_{BM}$, there must be a set $o_i$ in $O_{BM}$ that contains all the observations that can be perceived when the value of $v_i$ is modified. If the observations for $v_i$ depend on the particular environment, then $o_i$ should be specified when the specific knowledge is provided (in our navigation example, for the state variable $robot\_loc$, the set of observations would be $o_1=\{cell_1,cell_2,...\}$, because the robot measures its location in terms of cells).

\paragraph{}
$\boldsymbol{T_{BM}}$: Set of state transitions. For each action $a_i$ and each value in $w_j$ (where $v_j$ is the variable that $a_i$ can modify) a probability distribution over $w_j$ must be provided. Each element of a probability distribution is a state transition, that can be specified either as a transition with a list of particular values as $\langle w_{jk},a_i,w_{jl},p \rangle \in T_{BM}$, or defined by a neighborhood relation as $\langle a_i,N,q \rangle \in T_{BM}$, where $w_{jk},w_{jl} \in w_j$, $p$ and $q$ are transition probabilities, and $N$ is a binary relation defined over $w_j$, such that for every pair $(w_{jk},w_{jl}) \in N$, then $\langle w_{jk},a_i,w_{jl},q \rangle \in T_{BM}$ (\textit{e.g.}, the tuples $\langle right,current\_cell,0.2 \rangle$ and $\langle right,is\_at\_right,0.8 \rangle$ could be used to describe the transition distribution for the action $right$ for a complete environment, see Fig. \ref{fig:nav-t}).

\paragraph{}
$\boldsymbol{Z_{BM}}$: Set of observation transitions. For each action $a_i$ and each value in $w_j$ (where $v_j$ is the variable that $a_i$ can modify and $o_j$ is the set of observations for $v_j$) a probability distribution over $o_j$ must be provided. Each element of a probability distribution is an observation transition, that can be specified either as a transition with a list of particular values as $\langle w_{jk},a_i,o_{jl},p \rangle \in Z_{BM}$, or defined by a neighborhood relation as $\langle a_i,N,q \rangle \in Z_{BM}$, where $w_{jk} \in w_j$, $o_{jl} \in o_j$, $p$ and $q$ are transition probabilities, and $N$ is a binary relation defined over $w_j$ and $o_j$, such that for every pair $(w_{jk},o_{jl}) \in N$, then $\langle w_{jk},a_i,o_{jl},q \rangle \in Z_{BM}$ (\textit{e.g.}, to specify the observation distribution from Fig. \ref{fig:nav-o}, one could use nine tuples of the form $\langle right,current\_cell,0.6 \rangle$, $\langle right,is\_at\_right,0.05 \rangle$, $\langle right,is\_above,0.05 \rangle$, and so on).

\subsubsection{Domain dynamics}
The domain dynamics is defined by a collection of fluents and logical rules that describe a set of neighborhood relations, deterministic and non-deterministic causal laws, state constraints and executability conditions. The purpose of these fluents and rules is to describe, in a declarative programming language (in our case SPARC \cite{balai2013sparc}), the dynamics of the interactions between the robot and the environment, so that the architecture uses this description (along with the specification of each basic module) to build a POMDP that models these interactions (see Section \ref{sec:b-pomdp}).

\paragraph{}
\textbf{Neighborhood relations}: For each action $a_i$, a pair of set of neighborhood relations defined over $w_j$, and over $w_j$ and $o_j$, must be specified to model which pairs value-value and value-observation are neighbors with respect to action $a_i$ (\textit{i.e.}, which values and observations are reachable from certain value if $a_i$ is executed); where $w_j$ and $o_j$ are the sets of values and observations for the variable that $a_i$ modifies (\textit{e.g.}, recalling the mobile robot example, the relations $is\_at\_right(\#w_0,\#w_0)$ and $current\_cell(\#w_0,\#w_0)$ would describe the neighborhood relations for the action $right$, as they model the possibilities of the robot moving and drifting, respectively). In SPARC, binary relations are encoded as fluents that receive two arguments.

\paragraph{}
\textbf{Causal laws}: For each action, a collection of deterministic and non-deterministic rules must be defined to specify the state and observation transitions for that action, \textit{i.e.}, the effects actions provoke. For example, in the navigation scenario, the rule $robot\_loc(X,I), \: is\_at\_right(Y,X), \: right \rightarrow robot\_loc(Y,I+1)$ would describe the deterministic effect for the action $right$. On  the other hand, Eq. (\ref{eq:cl-non-det}) would specify the non-deterministic causal law, for the same action, along with the fluent $right\_possible\_outcome(\cdot,\cdot)$.

\begin{equation}\label{eq:cl-non-det}
    \begin{aligned}
        &robot\_loc(X,I), \: right \rightarrow \\
        &\{robot\_loc(Y,I+1) \mid right\_possible\_outcome(Y,I) \}
    \end{aligned}
\end{equation}

\paragraph{}
The second line in Eq. (\ref{eq:cl-non-det}) specifies the set of possible locations for the robot at the next step of time (\textit{i.e.}, $I+1$), considering that the action $right$ was executed and the definition of the fluent $right\_possible\_outcome(\cdot,\cdot)$. This fluent must be specified, by the designer, as a collection of deterministic rules. For instance, Eqs. (\ref{eq:cl-non-det-2}) and (\ref{eq:cl-non-det-3}) would describe the fluent for the non-deterministic effect of the action $right$, shown in Fig. \ref{fig:nav-t}.

\begin{equation}\label{eq:cl-non-det-2}
    \begin{aligned}
        &robot\_loc(X,I), \: current\_cell(X,Y) \rightarrow \\
        &right\_possible\_outcome(Y,I)
    \end{aligned}
\end{equation}

\begin{equation}\label{eq:cl-non-det-3}
    \begin{aligned}
        &robot\_loc(X,I), \: is\_at\_right(Y,X) \rightarrow \\
        &right\_possible\_outcome(X,I)
    \end{aligned}
\end{equation}

\paragraph{}
Thus, with non-deterministic causal laws a designer can encode the stochastic effect of any action, by using as many rules (with the form of Eqs. (\ref{eq:cl-non-det-2}) and (\ref{eq:cl-non-det-3})) as necessary. Moreover, in order for the system to include a possible effect (according to a non-deterministic causal law) in the bottom POMDP (see Section \ref{sec:b-pomdp}), the probability for such transition must be specified in the set $T_{BM}$ of a basic module.

\paragraph{}
\textbf{State constraints}: A collection of deterministic rules that model situations whose occurrence is impossible, given the nature of the environment (\textit{e.g.}, in the navigation example, the rule $robot\_loc(X,I),X \neq Y \rightarrow \neg robot\_loc(Y,I)$ would model the fact that a robot cannot be at several locations at the same time).

\paragraph{}
\textbf{Executability conditions}: For each action, a collection of deterministic rules must be provided to specify the situations in which an action should not be performed (\textit{e.g.}, in the navigation example, $robot\_loc(X,I),blocked\_above(X) \rightarrow \neg up$ would tell the system that $up$ should not be executed if the cell from above it is not free, which would apply for cell 1 in Fig. \ref{fig:example}, as there is no cell above it).\footnote{For more detail on how to define causal laws, state constraints and executability conditions in SPARC, read \cite{gelfond2014knowledge}.}

\paragraph{}
Furthermore, it is worth mentioning that if the designer forgets to add a state constraint, or an executability condition, the system can still work. However, the more constraints and conditions are included in the dynamics description, the less time it will take to build the bottom POMDP (see Section \ref{sec:b-pomdp}), as the system will evaluate a smaller number of state transitions.


\subsubsection{Hierarchical Function}
The system requires a function that describes a hierarchy, whose leaf nodes are the values for one of the state variables, while internal nodes are abstract values that the designer must provide in the specific knowledge. Such function should map a value to its parent within the hierarchy (in Fig. \ref{fig:example}, a function $is\_in(X)$ would map a location to a more abstract location that contains it, \textit{e.g.}, $is\_in(R2) = B1$, etc.).


\subsection{Specific knowledge}
The specific knowledge describes the particular environment in which the robot will operate. This description consists of a list of objects that characterize the whole scenario (concrete values), the objects in which the state space can be abstracted (abstract values), the pairs of objects that are neighbors by some action (neighborhood pairs) and the pairs of objects that exist in the hierarchical function (hierarchical function pairs). Moreover, for a robot to operate in different environments, the general knowledge can be reused and only the specific knowledge needs to be encoded for each environment.

\subsubsection{Concrete values}
For state variables whose sets of values (or observations) were not specified in the basic module definition, the designer must provide the values (or observations) that represent the particular environment in which the robot will operate. For instance, for the mobile robot of our example, the list of cells the robot can perceive (observations) and be at (state variable values) would be specified in this part of the KB (\textit{e.g.}, in Fig. \ref{fig:example}, this would mean $\{C1,C2,...,C12\}$).

\subsubsection{Abstract values}
The set $AW$ must contain the values that are a less detailed version of the values over which the hierarchical function is defined (\textit{e.g.}, in the environment from Fig. \ref{fig:example}, this set would be specified by all the sections, rooms and buildings that constitute the environment, \textit{i.e.}, $AW = \{S1,S2,...,S6,R1,R2,R3,B1,B2\}$).

\subsubsection{Neighborhood pairs}
For each neighborhood relation, the designer must provide the value-value and value-observation pairs that are true within the particular environment in which the robot will operate (\textit{e.g.}, in Fig. \ref{fig:example}, the pairs for relation $is\_above$ would be every pair of cells that are adjacent and aligned vertically, \textit{i.e.}, $\{(C1,C3),(C2,C4),...\}$).

\subsubsection{Hierarchical function pairs}
Let $w_i$ be the set of values (for state variable $v_i$) over which the hierarchical function will be defined, and $F : (w_i \cup AW) \rightarrow (w_i \cup AW)$ be the hierarchical function. Then, the designer must provide a subset of $(w_i \cup AW) \times (w_i \cup AW)$ that describes $F$, \textit{i.e.}, a function that maps every element in $(w_i \cup AW)$ to its parent value in the particular environment in which the robot will operate (\textit{e.g.}, in Fig. \ref{fig:example}, the hierarchical function $F = is\_in$ would be specified by $\{(C3,S2),...,(R2,B1),...\}$).


\section{Architecture initialization} \label{sec:ai}
After the general and specific knowledge have been specified, the architecture uses the domain dynamics, and the set of concrete values, to build a stochastic transition diagram. The sets of state and observation transitions are used to assign probabilities in the transition diagram, leading to the definition of the bottom POMDP. Then, a recursive definition of abstract actions is employed to build a hierarchy of POMDPs with the bottom POMDP and the hierarchical function, in a bottom-up approach.

\subsection{Construction of the bottom POMDP}\label{sec:b-pomdp}
Let a POMDP be defined by a tuple $M = \langle S,A,\Phi,R,O,\Omega,B_0 \rangle$ where $S$, $A$, and $O$ are the sets of states, actions and observations, respectively, $\Phi$, $\Omega$ and $R$ are the transition, observation and reward function, respectively, and $B_0$ the initial belief distribution. The purpose of the bottom POMDP is to describe the dynamics of the environment at the bottom level of the hierarchy. Thus, the basic modules, domain dynamics and specific knowledge are employed to define all the parameters of $M$, with the exception of $R$ and $B_0$. Each of the parameters of the bottom POMDP, which we will refer to as $BP = \langle S,A,\Phi,O,\Omega \rangle$, are defined as follows.

\paragraph{}
$\boldsymbol{S}$: The set of states is the cross product of the $n$ sets of values $\{w_1,...,w_n\}$, where $w_i$ is the set of values for a state variable $v_i$, defined in a basic module. Each element in $S$ is an $n$-tuple, where the \textit{i-th} element of a state $s$, \textit{i.e.} $s[i]$, is a value that $v_i$ can take. For instance, for the mobile robot and the environment from Fig. \ref{fig:example}, $S$ would be defined by set of all cells, since there is only one state variable.

\paragraph{}
$\boldsymbol{A}$: The set of actions is the union of all the sets of actions $A_{BM}$, where $A_{BM}$ is the set of actions of a basic module. For the mobile robot from Fig. \ref{fig:nav-a}, $A$ would be defined by $\{up,down,left,right\}$.

\paragraph{}
$\boldsymbol{O}$: The set of observations is the union of all the sets of observations $o_i$, where $o_i$ is the set of observations for a state variable $v_i$ defined in a basic module. For the mobile robot from Fig. \ref{fig:nav-ato} and the environment from Fig. \ref{fig:example}, $O$ would be defined by set of all twelve cells, since the robot perceives in terms of cells.

\paragraph{}
$\boldsymbol{\Phi}$: The transition function is the set of every tuple $\langle s_i,a,s_j,p\rangle$ where $s_i,s_j \in S$, $a \in A$, and $p$ is the transition probability, such that there is a tuple $\langle s_i[k],a,s_j[k],p \rangle$ in $T_{BM}$, for some index $k$. For the robot from Fig. \ref{fig:nav-t}, the transition distribution of every action would be specified by the pair of neighborhood relations that model the possibility of the robot drifting or moving towards its target cell.

\paragraph{}
$\boldsymbol{\Omega}$: The observation function is the set of every tuple $\langle s,a,o,p\rangle$ where $s \in S$, $a \in A$, $o \in O$ and $p$ is the observation probability, such that there is a tuple $\langle s[k],a,o,p \rangle $ in $Z_{BM}$ for some index $k$. For the robot shown in Fig. \ref{fig:nav-o}, the observation distribution of every action would be specified by the nine neighborhood relations that describe the $3 \times 3$ kernel.

\paragraph{}
Once the bottom POMDP has been built, it serves as input parameter for the construction of the hierarchy of POMDPs (see Algorithm \ref{alg:hie-act-const}).


\subsection{Construction of the hierarchy of POMDPs}\label{sec:con-aa}
The hierarchical function, defined in the KB, is used to build a hierarchical representation of $S$. This representation is employed, along with the $BP$, to build a hierarchy of POMDPs (each one representing an abstract action), that the architecture can later use to generate plans.


\subsubsection{State space tree}
The State space tree (SST) is a tree structure, where each node is an $n$-tuple and the set of leaf nodes is constituted by $S$ (the set of states of the $BP$). In the SST, the parent node ($n_{par}$) of a node $n_{chi}$ is defined by Eq. (\ref{eq:parent_rel}), for $i = 1,...,n$.
    \begin{equation}\label{eq:parent_rel}
        n_{par}[i] = \begin{cases}
        		n_{chi}[i] &\text{if $i \neq j$}\\
        		F(n_{chi}[i]) &\text{if $i = j$}
        \end{cases}
    \end{equation}
where $j$ is the index of the set of values over which the hierarchical function $F$ is defined and $(n_{chi}[j],F(n_{chi}[j])) \in F$ is a hierarchical function pair that is specified in the knowledge base, see Fig. \ref{fig:example} for an example of an SST. Moreover, internal nodes of the SST will be referred to as abstract states, and the children nodes of an abstract state as its children states.


\subsubsection{Neighbor states}
Let $s_i,s_j \in S$ be two states from the $BP$, such that there is an action $a \in A$, whose (at least) one of its neighborhood relations has the pair $\langle s_i[k],s_j[k] \rangle$ for some index $k$, then $s_i$ and $s_j$ are neighbor states. Furthermore, if in the SST a pair of nodes $n_a$ and $n_b$ are neighbor states and have different parents, then their parents are also neighbor states (\textit{e.g.}, in Fig. \ref{fig:example}, $s1$ and $s3$ are neighbor states because $C2$ and $C5$ are neighbors).


\subsubsection{Abstract actions as POMDPs}\label{subsub:aa-as-p}
Let $s_i^d$ and $s_j^d$ be two abstract neighbor states located at height $d$ in the SST and $C(s)$ the set of children states of $s$ in the SST. An abstract action $a_{ij}^d$, designed to transit from state $s_i^d$ to $s_j^d$ is modeled as a POMDP using the following formulation.

\begin{figure}[t!]
    \centering
    \includegraphics[width=\linewidth]{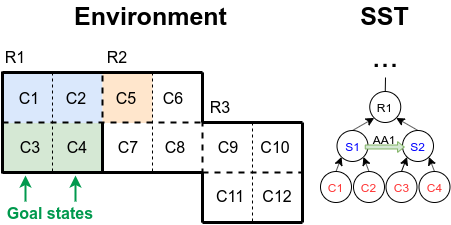}
    \caption{State space for the abstract action $AA1$, designed to transit from section $S1$ to $S2$ (based on the SST shown in Fig. \ref{fig:example}). The blue cells are non-goal states, the green cells represent the goal states, while the orange cell is a state the agent should avoid transiting to. Every white cell is not part of the local state space and, therefore, is represented by the $extra$ state.}
    \label{fig:aa-states}
\end{figure}

\paragraph{}
$\boldsymbol{S_{ij}^{d}}$: Set of states from the immediate lower level in the SST that are relevant for abstract action $a_{ij}^d$.
\begin{equation}\label{eq:def-states-aa}
    \begin{aligned}
        S_{ij}^{d} = & C(s_i^d) \\
        & \bigcup \{s \mid s \notin C(s_i^d), \exists s_k \in C(s_i^d) , neighbors(s,s_k) \} \\
        & \bigcup \{ extra, absb\_g, absb\_{ng} \}
    \end{aligned}
\end{equation}
That is, $S_{ij}^{d}$ is constituted by the children states of $s_i^d$ (blue cells in Fig. \ref{fig:aa-states}), any neighbor to its children states (orange and green cells in Fig. \ref{fig:aa-states}) and by the special states $extra$, $absb\_g$ and $absb\_{ng}$. The purpose of the $extra$ state is to represent the portion of the state space that is not modeled in $S_{ij}^{d}$ (white cells in Fig. \ref{fig:aa-states}). On the other hand, the objective of the absorbent states $absb\_g$ and $absb\_{ng}$ is to represent the end of the policy execution, as a consequence of reaching its goal state or a state that is not in $S_{ij}^{d}$, respectively.

\paragraph{}
$\boldsymbol{A_{ij}^{d}}$: Set of actions that can be executed at the immediate lower level in the SST and have a probability greater than 0 of transiting between a pair of states in $S_{ij}^{d}$, as well as a special action.
\begin{equation}\label{eq:def-actions-aa}
    \begin{aligned}
        A_{ij}^{d} = & \{ a \mid \exists s_k , s_l \in S_{ij}^d , \Phi^{d+1}(s_k , a , s_l) > 0 \} \\
        & \bigcup \{ terminate \}
    \end{aligned}
\end{equation}
With regards to the special action $terminate$, this action is added so that, during the operation stage of the architecture (AO), the policy (obtained from the POMDP $a_{ij}^d$) has a way to indicate that it believes it has reached its goal state, or the $extra$ state. For instance, in Fig. \ref{fig:example}, the abstract actions built to transit between sections, include $terminate$ and the concrete actions (\textit{i.e.}, actions from the BP) in their action set $A_{ij}^{d}$.

\paragraph{}
$\boldsymbol{O_{ij}^{d}}$: Set of observations from the immediate lower level in the SST that have a probability greater than 0 of being perceived after reaching a state in $S_{ij}^{d}$ with an action in $A_{ij}^{d}$, as well as two special observations.
\begin{equation}\label{eq:def-observations-aa}
    \begin{aligned}
        O_{ij}^{d} = & \{ o \mid \exists s \in S_{ij}^{d}, \exists a \in A_{ij}^{d},\Omega^{d+1}(s,a,o) > 0 \} \\
        & \bigcup \{ none, extra\}
    \end{aligned}
\end{equation}
Also, because the $terminate$ action does not have an effect in the environment, and does not generate observations, the purpose of the special observation $none$ is to represent such lack of information the system can expect after executing $terminate$. Whereas the $extra$ observation, similar to the $extra$ state, will be returned to the system whenever a non-modeled observation is perceived. For instance, for the abstract action in Fig. \ref{fig:aa-states}, $O_{ij}^{d}$ would include those observations that can be perceived from cells 1, 2, 3, 4 and 5, according to the kernel from Fig. \ref{fig:nav-o} that models the observation distribution of every concrete action.

\paragraph{}
$\boldsymbol{\Phi_{ij}^{d}}$: For every pair of states in $S_{ij}^{d}$ (other than $extra$, $absb\_g$ and $absb\_{ng}$) the transition function simply takes the probability values defined in $\Phi^{d+1}$ (transition function of the immediate lower level in the SST) for every non-special action in $A_{ij}^{d}$.
\begin{equation}
    \begin{aligned}
        & \forall s_k, s_l \in S_{ij}^{d} \setminus \{extra,absb\_g,absb\_{ng}\},\\
        & \forall a \in A_{ij}^{d} \setminus \{terminate\},\\
        & \Phi_{ij}^{d}(s_k,a,s_l) = \Phi^{d+1}(s_k,a,s_l)
    \end{aligned}
\end{equation}
However, since there may be some state-action pairs that have a transition distribution with ending states that are not in $S_{ij}^{d}$, the ending state for those transitions is substituted by the $extra$ state, as Eq. (\ref{eq:aa-t-2}) specifies.
\begin{equation}\label{eq:aa-t-2}
    \begin{aligned}
        & \forall s_k \in S_{ij}^{d} \setminus \{extra,absb\_g,absb\_{ng}\},\\
        & \forall a \in A_{ij}^{d} \setminus \{terminate\},\\
        & S_e = \{s \mid s \notin S_{ij}^{d}, s \in S^{d+1}\},\\
        & \Phi_{ij}^{d}(s_k,a,extra) = \sum_{s \in S_e} \Phi^{d+1}(s_k,a,s)
    \end{aligned}
\end{equation}
With regards to executing actions from the $extra$ state, the model is told that they do not have any effect, despite that this may not be true, as specified in Eq. (\ref{eq:aa-t-3}). That is, since the $extra$ state only informs the model that the agent is out of its local state space, but does not provide information of its actual state, the best the agent can do from the $extra$ state (as we will see in the definition of the reward function) is to execute $terminate$, \textit{i.e.}, end the execution of the current abstract action and return the control to the system, so that it invokes other abstract action that is more appropriate given the current belief state (see Section \ref{sec:exec-pols} for more detail on the execution of abstract actions).
\begin{equation}\label{eq:aa-t-3}
    \begin{aligned}
        & \forall a \in A_{ij}^{d} \setminus \{terminate\}, \\
        & \Phi_{ij}^{d}(extra,a,extra) = 1.0
    \end{aligned}
\end{equation}
Also, the distribution for executing any action from any of two absorbent states has no effect, as Eq. (\ref{eq:aa-t-4}) specifies.
\begin{equation}\label{eq:aa-t-4}
    \begin{aligned}
        & \forall a \in A_{ij}^{d}, \forall s \in \{absb\_g,absb\_ng\} \\
        & \Phi_{ij}^{d}(s,a,s) = 1.0
    \end{aligned}
\end{equation}
Regarding the distributions for the $terminate$ action and non-absorbent states, these are specified by Eq. (\ref{eq:aa-t-5}) and Eq. (\ref{eq:aa-t-6}).
\begin{equation}\label{eq:aa-t-5}
    \begin{aligned}
        & \forall s \in S_{ij}^{d} \bigcap C(s_j^d), \\
        & \Phi_{ij}^{d}(s,terminate,absb\_g) = 1.0
    \end{aligned}
\end{equation}

\begin{equation}\label{eq:aa-t-6}
    \begin{aligned}
        & \forall s \in S_{ij}^{d} \setminus (\{absb\_g,absb\_{ng}\} \bigcup C(s_j^d)), \\
        & \Phi_{ij}^{d}(s,terminate,absb\_ng) = 1.0
    \end{aligned}
\end{equation}
Where $C(s_j^d)$ is the set of children states of the target state for abstract action $a_{ij}^d$ (green cells in Fig. \ref{fig:aa-states}). That is, the model can expect to reach a different absorbent state by executing $terminate$ from a goal state (a child of $s_j^d$) than from a non-goal state (any other non-absorbent state, \textit{e.g.}, the orange, blue and white cells in Fig. \ref{fig:aa-states}).

\paragraph{}
$\boldsymbol{\Omega_{ij}^{d}}$: The observation function takes the probability value, defined in $\Omega^{d+1}$, for every element in $S_{ij}^{d}$, $A_{ij}^{d}$, and $O_{ij}^{d}$, excluding the special components, as Eq. (\ref{eq:aa-o-1}) illustrates.
\begin{equation}\label{eq:aa-o-1}
    \begin{aligned}
        & \forall s \in S_{ij}^{d} \setminus \{extra,absb\_g,absb\_{ng}\}, \\
        & \forall a \in A_{ij}^{d} \setminus \{terminate\}, \\
        & \forall o \in O_{ij}^{d} \setminus \{extra,none\}, \\
        & \Omega_{ij}^{d}(s,a,o) = \Omega^{d+1}(s,a,o)
    \end{aligned}
\end{equation}
The distribution for the $terminate$ action is specified by Eq. (\ref{eq:aa-o-2}).
\begin{equation}\label{eq:aa-o-2}
    \begin{aligned}
        & \forall s \in S_{ij}^{d}, \\
        & \Omega_{ij}^{d}(s,terminate,none) = 1.0
    \end{aligned}
\end{equation}
With regards to the distribution of the non-terminal actions and special states, these are defined by Eq. (\ref{eq:aa-o-3})  and Eq. (\ref{eq:aa-o-4}).
\begin{equation}\label{eq:aa-o-3}
    \begin{aligned}
        & \forall s \in \{absb\_g,absb\_{ng}\}, \forall a \in A_{ij}^{d} \setminus \{terminate\}, \\
        & \Omega_{ij}^{d}(s,a,none) = 1.0
    \end{aligned}
\end{equation}

\begin{equation}\label{eq:aa-o-4}
    \begin{aligned}
        & \forall a \in A_{ij}^{d} \setminus \{terminate\}, \\
        & \Omega_{ij}^{d}(extra,a,extra) = 1.0
    \end{aligned}
\end{equation}
That is, with Eq. (\ref{eq:aa-o-3}) the model should not expect to gather any useful observation once it has reached an absorbent state, while Eq. (\ref{eq:aa-o-4}) indicates that, whenever the agent perceives the $extra$ observation, it can be sure to be in the $extra$ state, since such observation is returned when a non-modeled observation is perceived, which can only be perceived from states that are not in $S_{ij}^{d}$, \textit{i.e.}, the those that are represented by the $extra$ state.

\paragraph{}
$\boldsymbol{R_{ij}^{d}}$: Since the purpose of the abstract action $a_{ij}^d$ is to reach $s_j^d$ from $s_i^d$, the reward function is designed based on two principles: 1) that it should reach $s_j^d$ as soon as possible and 2) to stop its execution if it believes to be outside of its local state space ($S_{ij}^{d}$). These principles are modeled by the following equations.
\begin{equation}\label{eq:aa-r-1-2}
    \begin{aligned}
        & \forall s \in S_{ij}^{d} \setminus \{ absb\_g, absb\_ng \}, \\
        & R_{ij}^{d}(s,terminate,\cdot) = \begin{cases}
    		\Re^{-} &\text{if $s \in C(s_i^d)$}\\
    		\Re^{+} &\text{otherwise}
        \end{cases}
    \end{aligned}
\end{equation}
where $\Re$ is a large scalar value. From Eq. (\ref{eq:aa-r-1-2}) we model the fact that the agent should only terminate the policy execution from states outside of $C(s_i^d)$ (\textit{e.g.}, in Fig. \ref{fig:aa-states}, cells that are not blue), whether if it is from its goal states (children states of $s_j^d$), from a children of other neighbor state of $s_i^d$, or any other region of the state space (the $extra$ state).
\begin{equation}\label{eq:aa-r-3}
    \forall s \in S_{ij}^{d} \setminus (C(s_i^d) \bigcup C(s_j^d)), \: R_{ij}^{d}(\cdot,\cdot,s) = \Re^{-}
\end{equation}

\begin{equation}\label{eq:aa-r-4}
    R_{ij}^{d}(\cdot,\cdot,extra) = \Re^{-}
\end{equation}
With Eq. (\ref{eq:aa-r-3}) and Eq. (\ref{eq:aa-r-4}) the agent is encouraged to avoid transiting to any other neighbor state of $s_i^d$ that is not $s_j^d$, or to the $extra$ state (\textit{e.g.}, the orange and white cells, respectively, in Fig. \ref{fig:aa-states}).
\begin{equation}\label{eq:aa-r-5}
    \forall a \in A_{ij}^{d} \setminus \{terminate\}, \: R_{ij}^{d}(extra,a,\cdot) = \Re^{-}
\end{equation}
Equation (\ref{eq:aa-r-5}) models the fact that if the agent happens to reach the $extra$ state, even though that is not desirable, it should terminate the policy execution so that the architecture can invoke a more appropriate action given the current belief state.
\begin{equation}\label{eq:aa-r-6}
    R_{ij}^{d}(absb\_g,terminate,\cdot) = \Re^{+}
\end{equation}
Because the absorbent state $absb\_g$ can only be reached by executing $terminate$ from any child state of $s_j^d$ (\textit{e.g.}, the green cells in Fig. \ref{fig:aa-states}), with Eq. (\ref{eq:aa-r-6}) we invite the agent to end the policy execution as fast as possible from a goal state, so that it can keep invoking $terminate$ for the remaining steps of the episode from $absb\_g$, and retrieve a larger accumulated reward than from any other state.
\begin{equation}\label{eq:aa-r-7}
    \begin{aligned}
        & \forall s_k, s_l \in S_{ij}^{d} \setminus \{extra\}, \forall a \in A_{ij}^{d} \setminus \{terminate\}, \\
        & \: R_{ij}^{d}(s_k,a,s_l) = -1
    \end{aligned}
\end{equation}
For any transition that does not start nor end in the $extra$ state, by executing a non-terminate action, the agent will retrieve a uniform negative reward. In this way, Eq. (\ref{eq:aa-r-7}) encourages the agent to not wander around, given that actions have a cost. Furthermore, when the actions in $A_{ij}^{d}$ are abstract actions, the reward signal for some transitions (that were already specified by Eq. (\ref{eq:aa-r-7})) is overwritten by Eq. (\ref{eq:aa-r-8}).
\begin{equation}\label{eq:aa-r-8}
    \begin{aligned}
        & \forall s_k, s_l \in S_{ij}^{d} \setminus \{extra\}, \forall a_{mn} \in A_{ij}^{d} \setminus \{terminate\}, \\
        & R_{ij}^{d}(s_k,a_{mn},s_l) = \begin{cases}
    		-1 &\text{if $s_k = s_m$}\\
    		\Re^{-} &\text{if $s_k \neq s_m$}
        \end{cases}
    \end{aligned}
\end{equation}
where $a_{mn}$ is the abstract action designed to transit from $s_m$ to $s_n$. Thus, Eq. (\ref{eq:aa-r-8}) expresses the fact that the agent should not invoke abstract actions from states that are not the initial state by design (\textit{e.g.}, in Fig. \ref{fig:aa-states}, the initial state by design of the abstract action AA1 is the abstract state S1).


\paragraph{}
Once the tuple $\langle S_{ij}^d,A_{ij}^d,O_{ij}^d,\Phi_{ij}^d,\Omega_{ij}^d,R_{ij}^d \rangle$ for the local POMDP that models the abstract action $a_{ij}^d$ is fully defined, a POMDP solving algorithm is employed to compute its policy. However, in order to use $a_{ij}^d$ as an action in another POMDP, its transition and observation distributions must be defined, as well as the state and observation spaces it is associated with. These distributions and spaces are specified as follows.

\paragraph{}
$\boldsymbol{S^d}$: The state space at height $d$ is constituted by every node of the SST located at height $d$.

\paragraph{}
$\boldsymbol{O^d}$: The set of abstract observations at height $d$ is made up by one observation $o^d$ for each abstract state $s^d \in S^d$.

\paragraph{}
$\boldsymbol{\Phi^d}$: For an abstract action $a_{ij}^d$, designed to transition from state $s_{i}^d$ to $s_{j}^d$, the transition probability distribution defined over abstract states at height $d$ is given by the following equations.
\begin{equation}\label{eq:T-s0-reach}
    \begin{aligned}
        & \forall s_k^d \in (\{s \mid neighbors(s_i^d,s)\} \bigcup \{s_i^d\}),\\
        & \Phi^d(s_i^d,a_{ij}^d,s_k^d) = sim\_prob(s_k^d)
    \end{aligned}
\end{equation}

\begin{equation}\label{eq:T-s0-unreach}
    \begin{aligned}
        & \forall s_k^d \in (S^d \setminus (\{s \mid neighbors(s_i^d,s)\} \bigcup \{s_i^d\})),\\
        & \Phi^d(s_i^d,a_{ij}^d,s_k^d) = 0.0
    \end{aligned}
\end{equation}

\begin{equation}\label{eq:T-not-s0}
    \begin{aligned}
        & \forall s_k^d \in \{s \mid s \neq s_i^d\}, \\
        & \Phi^d(s_k^d,a_{ij}^d,s_k^d) = 1.0
    \end{aligned}
\end{equation}
where $sim\_prob(s)$ is a probability that is estimated by simulating the policy obtained after solving the POMDP that models $a_{ij}^d$. That is, let $M$ be the amount of times the policy for $a_{ij}^d$ is simulated, $SimCount(s)$ the amount of simulations that end at a child of abstract state $s$, and $Neig(s_i^d)$ the set of neighbor states of $s_i^d$, then, the probability of transiting from $s_i^d$ to one of its neighbors, or to $s_i^d$ itself, with action $a_{ij}^d$ is estimated by Eq. (\ref{eq:sim-prob}).
\begin{equation}
    \begin{aligned}\label{eq:sim-prob}
        & \forall s \in \{s_i^d\} \bigcup Neig(s_i^d),\\
        & sim\_prob(s) = \frac{SimCount(s)}{M}
    \end{aligned}
\end{equation}
In each simulation of the policy for $a_{ij}^d$, the initial state $s_0$ is randomly sampled from $S_{ij}^{d} \setminus \{extra,absb\_g,absb\_ng\}$ (following a uniform distribution), and in the initial belief distribution, $s_0$ has a probability of $1.0$. Then, actions are sampled from the policy for $a_{ij}^d$, while the resulting state and perceived observation are sampled from the transition and observation functions located at the immediate level below, \textit{i.e.}, $\Phi^{d+1}$ and $\Omega^{d+1}$. Finally, a simulation ends when the policy invokes the $terminate$ action from the final state $s_{final}$, then, the count of a state $s$ (\textit{i.e.}, $SimCount(s)$) increases by one if $s_{final} \in C(s)$.

\paragraph{}
Regarding the transition distribution, Eq. (\ref{eq:T-s0-reach}) assigns the probability for transitions that start in $s_i^d$ and end in a reachable state, Eq. (\ref{eq:T-s0-unreach}) for transitions that end in unreachable states, whereas Eq. (\ref{eq:T-not-s0}) expresses the fact that $a_{ij}^d$ can only change the state of the agent if it is invoked from $s_i^d$.

\paragraph{}
$\boldsymbol{\Omega^d}$: For an abstract action $a_{ij}^d$, designed to transit from state $s_{i}^d$ to $s_{j}^d$, the probability of perceiving an abstract observation after executing $a_{ij}^d$ is given by Eq. (\ref{eq:O-aa}).
\begin{equation}\label{eq:O-aa}
    \begin{aligned}
        & \forall s_k^d \in S^d,\\
        & \Omega^d(s_k^d,a_{ij}^d,o_k^d) = 1.0
    \end{aligned}
\end{equation}
where $o_k^d$ is the observation associated to the abstract state $s_k^d$.

\paragraph{}
The reason the observation distribution is defined without uncertainty is because the agent will not perceive abstract observations, only concrete ones (\textit{i.e.}, observations from the BP). During the operation stage, the agent does not need abstract observations to update the belief state vector at an internal level in the SST, but it rather synthesizes state probabilities upwards from the bottom of the SST (see Section \ref{sec:global-b} for more detail on the updates of belief distributions in the SST). Furthermore, as experimental results suggest, this simplification does not seem to affect negatively the capacity of the system to select actions during operation.

\begin{algorithm}
    \caption{Construction of hierarchy of actions}\label{alg:hie-act-const}
    \begin{algorithmic}[1]
        \Procedure{BuildHierarchyActions}{$BP$, $SST$, $M$}
	        \State $i \gets depth(SST)-1$
	        \State $H \gets [\:]$
	        \While{$i > 0$}
	        		\State $S \gets States(SST[i])$
	        		\State $O \gets GenerateObs(S)$
	        		\State $A \gets \emptyset$
	        		\State $\Phi \gets \emptyset$
	        		\State $\Omega \gets \emptyset$
	        		\ForAll{$s_0 \in S$}
	        			\ForAll{$s_1 \in NeighborStates(s_0)$}
	        				\State $a \gets BuildAA(s_0,s_1,H[len(H)-1])$
	        				\State $t$, $z \gets EstimateTZ(a,S,O,M)$
	        				\State $A \gets A \bigcup \{a\}$
	        				\State $\Phi \gets \Phi \bigcup t$
	        				\State $\Omega \gets \Omega \bigcup z$
	        			\EndFor
	        		\EndFor
	        		\State $Append(H,\langle S,A,O,\Phi,\Omega \rangle)$
	        		\State $i \gets i-1$
	        \EndWhile
	        \State \textbf{return} $H$
        \EndProcedure
    \end{algorithmic}
\end{algorithm}

\paragraph{}
In Algorithm \ref{alg:hie-act-const} the hierarchy of actions (modeled as POMDPs) is built. It receives as input parameters the bottom POMDP ($BP$), the hierarchical representation of the state space ($SST$) and the amount of times abstract actions should be simulated to estimate their parameters ($M$). For each internal level in the SST, a tuple $\langle S,A,O,\Phi,\Omega \rangle$ is specified, in a bottom-up way. At every internal level in the SST, the formulation presented in section \ref{subsub:aa-as-p} is employed (lines 12 and 13) to build an abstract action for every ordered pair of abstract states that are neighbors, that is, $a_{ij}^d \neq a_{ji}^d$.


\section{Architecture operation} \label{sec:ao}
During its operation phase, the agent is ready to receive task requests, which must be passed as a goal state, which is an $n$-tuple that specifies a value for each one of the $n$ state variables (defined in the KB). Then, a \textit{hierarchical policy} is built and executed in a top-down way to gradually bring the agent to the goal state.

\subsection{Construction of a hierarchical policy}\label{sub:}
A hierarchical policy (HP) is a vector of POMDP policies (one for each level in the SST, except for the level of the root node) which we will refer to as Local Policies (LP). To build an HP, it is necessary to represent the goal state at every level within the SST. Thus, we define the \textit{hierarchical state}, which is a path from the root of the SST to the goal state (\textit{e.g.}, in Fig. \ref{fig:example}, the hierarchical state of cell 12 is $[root,B2,R3,S6,C12]$). In Algorithm \ref{alg:hie-pol} is shown a summary of the procedure employed to build an HP.

\begin{algorithm}
    \caption{Construction of hierarchical policy}\label{alg:hie-pol}
    \begin{algorithmic}[1]
        \Procedure{HierarchicalPolicy}{$G$, $H$, $SST$}
            \State $G^H \gets HierarchicalState(G,SST)$
            \State $\Pi^H \gets [\:]$
            \State $parent\_node \gets G^H[0]$
            \State $i \gets 1$
            \While{$i < len(G^H)$}
                \State $j \gets len(H)-i$
                \State $S \gets Children(parent\_node)$
                \State $S \gets S \bigcup \{ ps \mid ps \notin S, \exists s \in S, neig(ps,s) \}$
                \State $S \gets S \bigcup \{ absb\_g, absb\_ng \}$
                \State $A \gets RelevantActions(S,H[j][1])$
                \State $A \gets A \bigcup \{ terminate \}$
                \State $O \gets RelevantObservations(S,H[j][2])$
                \State $O \gets O \bigcup \{ none \}$
                \If {$i > 1$}
                    \State $S \gets S \bigcup \{ extra \}$
                    \State $O \gets O \bigcup \{ extra \}$
                    \State $A \gets A \bigcup \{ help \}$
                \EndIf
                \State $\Phi \gets TransitionSpec(S,A,G^H[i],H[j][3])$
                \State $\Omega \gets ObservationSpec(S,A,O,H[j][4])$
                \State $R \gets GoalBasedReward(S,A,G^H[i])$
                \State $\pi \gets SolvePolicy(S,A,O,\Phi,\Omega,R)$
                \State $Append(\Pi^H,\pi)$
                \State $parent\_node \gets G^H[i]$
                \State $i \gets i+1$
            \EndWhile
            \State \textbf{return} $\Pi^H$
        \EndProcedure
    \end{algorithmic}
\end{algorithm}

\paragraph{}
Algorithm \ref{alg:hie-pol} receives as parameters the goal state ($G$), the hierarchy of actions ($H$) and the state space tree ($SST$). For every node in the hierarchical state of $G$ (except for the root of the SST) an LP is built. In the POMDP of the LP designed to reach $G^H[i]$ at height $i$ in the SST, $S$ is composed by the absorbent states, the goal state $G^H[i]$, the states that are siblings of $G^H[i]$ in the SST (states that share their parent node with $G^H[i]$) and any other state that is neighbor to any of them and does not share parent with them. The sets $A$ and $O$ are defined following the same criteria used in the construction of abstract actions (see Eq. (\ref{eq:def-actions-aa}) and Eq. (\ref{eq:def-observations-aa})), with the difference that, in Algorithm \ref{alg:hie-pol}, the height from which actions and observations are being gathered is expressed by $i$, instead of $d+1$. Another difference is that, in Algorithm \ref{alg:hie-pol}, the $extra$ observation is added in certain cases, contrary to the construction of abstract actions.

\paragraph{}
In Algorithm \ref{alg:hie-pol}, since at height $1$ (immediately below the root of the SST) the set $S$ of the LP will consist of every state that is in that height, the $extra$ state and observation are not needed (see the top LP in the HP shown in Fig. \ref{fig:h-pol}). Also, the special action $help$ is not required, since its purpose is to return the control to the LP from the level above it in the SST (see Section \ref{sec:exec-pols}), \textit{i.e.}, no LP will be built above level $1$. Regarding the transition and observation distributions, the criteria used for the specification of $\Phi_{ij}^{d}$ and $\Omega_{ij}^{d}$ in the abstract actions is reused for LPs, however, the following modifications are made so that an LP incorporates the action $help$.
\begin{equation}\label{eq:help-T-add-1}
    \begin{aligned}
        & \forall s \in S \setminus \{absb\_g\},\\
        & \Phi(s,help,absb\_ng) = 1.0
    \end{aligned}
\end{equation}

\begin{equation}\label{eq:help-T-add-2}
    \begin{aligned}
        & \Phi(absb\_g,help,absb\_g) = 1.0
    \end{aligned}
\end{equation}

\begin{equation}\label{eq:help-O-add}
    \begin{aligned}
        & \forall s \in S,\\
        & \Omega(s,help,none) = 1.0
    \end{aligned}
\end{equation}

\begin{equation}\label{eq:help-T-mod-1}
    \begin{aligned}
        & \Phi(extra,terminate,extra) = 1.0
    \end{aligned}
\end{equation}

\begin{equation}\label{eq:help-T-mod-2}
    \begin{aligned}
        & \Phi(G^H[i],terminate,absb\_g) = 1.0
    \end{aligned}
\end{equation}

\begin{equation}\label{eq:help-T-mod-3}
    \begin{aligned}
        & \forall s \in S \setminus \{absb\_g,absb\_ng\}, s \neq G^H[i]\\
        & \Phi(s,terminate,absb\_ng) = 1.0
    \end{aligned}
\end{equation}
Equations (\ref{eq:help-T-add-1}), (\ref{eq:help-T-add-2}) and (\ref{eq:help-O-add}) are added to the formulation to specify the distributions for the $help$ action, whereas, Eqs. (\ref{eq:help-T-mod-1}), (\ref{eq:help-T-mod-2}) and (\ref{eq:help-T-mod-3}) are a modification of Eqs. (\ref{eq:aa-t-3}), (\ref{eq:aa-t-5}) and (\ref{eq:aa-t-6}), respectively. That is, Eq. (\ref{eq:help-T-mod-1}) models the fact that, in an LP, it is the $help$ action that takes the agent out of the $extra$ state, instead of $terminate$. On the other hand, Eqs. (\ref{eq:help-T-mod-2}) and (\ref{eq:help-T-mod-3}) modify the formulation because, in an LP, the goal state is $G^H[i]$, rather than a set of neighbor states.

\paragraph{}
With regards to the reward function of an LP, Algorithm \ref{alg:hie-pol} also reuses the formulation employed to define the reward function for abstract actions, with the following modifications due to the addition of action $help$, and the difference in the definition of what a goal state is.
\begin{equation}\label{eq:help-R-add-1}
    \begin{aligned}
        & \forall s \in S, \\
        & R_{ij}^{d}(s,help,\cdot) = \begin{cases}
    		\Re^{+} &\text{if $s = extra$}\\
    		\Re^{-} &\text{if $s \neq extra$}
        \end{cases}
    \end{aligned}
\end{equation}

\begin{equation}\label{eq:help-R-add-2}
    \forall a \in A \setminus \{ help \}, \: R_{ij}^{d}(extra,a,\cdot) = \Re^{-}
\end{equation}

\begin{equation}\label{eq:help-R-mod-1}
    \begin{aligned}
        & \forall s \in S \setminus \{ absb\_g \}, \\
        & R_{ij}^{d}(s,terminate,\cdot) = \begin{cases}
    		\Re^{+} &\text{if $s = G^H[i]$}\\
    		\Re^{-} &\text{if $s \neq G^H[i]$}
        \end{cases}
    \end{aligned}
\end{equation}
Equations (\ref{eq:help-R-add-1}) and (\ref{eq:help-R-add-2}) motivate the agent to execute the $help$ action only from the $extra$ state, \textit{i.e.}, when the agent derails from the local state space of the LP. Equation (\ref{eq:help-R-mod-1}) is a modification of Eq. (\ref{eq:aa-r-1-2}), since in an LP the goal state is specified as $G^H[i]$.

\subsection{Execution of policies}
The execution of an HP is performed by two procedures that interleave the control of the execution process. Algorithm \ref{alg:pomdp-pol-exec} is in charge of executing POMDP policies (such LPs and abstract actions), while Algorithm \ref{alg:hie-pol-exec} determines the order in which LPs should be executed to take the agent to the goal state. Moreover, in order for these algorithms to keep track of the changes each action causes (at all levels in the SST), we define a multi-resolution representation of the belief state of the agent, which is described in Section \ref{sec:global-b}, whereas the algorithms that are responsible for the execution of POMDP policies, and hierarchical policies, are presented in Section \ref{sec:exec-pols}.

\subsubsection{Global belief}\label{sec:global-b}
The global belief (GB) is a tree, whose structure is identical to the SST, however, each node in the GB consists of an ordered pair $gb=[s,p]$, where $s$ is the correspondent node in the SST, and $p \in \mathbb{R}$ represents the belief probability of $s$. The GB represents the belief state distributions at every level of the SST, and is employed by Algorithm \ref{alg:pomdp-pol-exec} so that a POMDP policy (whether it is an abstract action or an LP from a hierarchical policy) samples actions during its execution. Algorithm \ref{alg:update-gb} illustrates the procedure followed to update the GB based on the last performed action, and the last perceived observation, at the bottom of the SST.

\begin{algorithm}
    \caption{Update of the global belief}\label{alg:update-gb}
    \begin{algorithmic}[1]
        \Procedure{UpdateGlobalBelief}{$B$, $a$, $z$}
            \State $UpdateBottomLevel(B,a,z)$
            \State $i \gets getHeight(B) - 1$
            \For{$i > 0$}
                \ForAll{$gb$ at level $i$ in $B$}
                    \State $gb[1] \gets 0.0$
                    \State $children \gets C(gb)$
                    \ForAll{$c$ in $children$}
                        \State $gb[1] \gets gb[1] + c[1]$
                    \EndFor
                \EndFor
                \State $i \gets i - 1$
            \EndFor
            \State \textbf{return} $B$
        \EndProcedure
    \end{algorithmic}
\end{algorithm}

\paragraph{}
Algorithm \ref{alg:update-gb} takes as parameters the GB to be updated ($B$), a concrete action ($a$) and a concrete observation ($z$). First, the belief distribution for the states at the bottom of the SST is updated with $a$ and $z$ (line $2$). Then, the belief probabilities are synthesized upwards. That is, the probability of an internal node $gb$, in $B$, is the sum of the probabilities of its children nodes $C(gb)$.

\subsubsection{Execution of a hierarchical policy}\label{sec:exec-pols}
Given that a hierarchical policy is constituted by a list of LPs (that are POMDP policies), whose actions may also be policies (abstract actions), we first introduce the execution of POMDP policies (upon which the execution of the HP takes place), followed by the description of the execution of a HP.

\begin{figure}[t!]
    \centering
    \includegraphics[width=\linewidth]{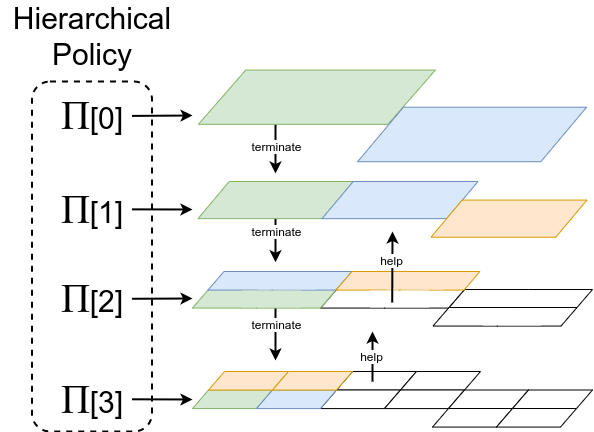}
    \caption{Hierarchical policy built in the environment from Fig. \ref{fig:example}. At each level in the SST (except level 0), an LP attempts to take the agent to the goal state in that level. When policy $\Pi[i]$ believes that it has reached its goal, it will invoke the $terminate$ action and pass the control to $\Pi[i+1]$. However, if $\Pi[i]$ believes that the agent has transited to the $extra$ state, it will invoke the $help$ action to return the control to $\Pi[i-1]$. The local state space of each LP, similar to the abstract action in Fig. \ref{fig:aa-states}, consists of non-goal (blue) and goal (green) states, as well of states that the agent should avoid transiting to (orange) and  the $extra$ state (white).}
    \label{fig:h-pol}
\end{figure}

\begin{algorithm}
    \caption{Execution of a POMDP policy}\label{alg:pomdp-pol-exec}
    \begin{algorithmic}[1]
        \Procedure{ExecutePolicy}{$\pi$, $B$, $d$}
            \State $S_{Loc} \gets getStateSpace(\pi)$
            \State $b \gets mapBeliefToLocal(S_{Loc},B,d)$
            \State $\pi_E \gets entropyWeight(\pi,S_{Loc},B,d)$
            \State $a \gets getAction(\pi_E,b)$
            \While{$a \neq terminate$ and $a \neq help$}
                \If{$a$ is concrete}
                    \State $z \gets executeAction(a)$
                    \State $B \gets updateGlobalBelief(B,a,z)$
                \Else
                    \State $[a,B] \gets ExecutePolicy(a,B,d+1)$
                \EndIf
            \State $b \gets mapBeliefToLocal(S_{Loc},B,d)$
            \State $\pi_E \gets entropyWeight(\pi,S_{Loc},B,d)$
            \State $a \gets getAction(\pi_E,b)$
            \EndWhile
            \State \textbf{return} $[a,B]$
        \EndProcedure
    \end{algorithmic}
\end{algorithm}

\paragraph{}
Algorithm \ref{alg:pomdp-pol-exec} takes as input parameters the policy to be executed ($\pi$), the current GB ($B$) and the height in the SST ($d$) at which the execution of the policy will take place. First, the state space of $\pi$ is stored in $S_{Loc}$, then a belief state distribution $b$ is built for the local state space $S_{Loc}$, based on the current GB (line 3). The belief probability for each state in $S_{Loc}$ is given by Eqs. (\ref{eq:map-b-1}), (\ref{eq:map-b-2}) and (\ref{eq:map-b-3}).

\begin{equation}\label{eq:map-b-1}
    \begin{aligned}
        & \forall gb_{Loc} \in \{gb \mid gb \in nodesAtHeight(B,d), gb[0] \in S_{Loc}\} \\
        & b(gb_{Loc}[0]) = gb_{Loc}[1]
    \end{aligned}
\end{equation}

\begin{equation}\label{eq:map-b-2}
    \begin{aligned}
        GB_{extra} = &\{gb \mid gb \in nodesAtHeight(B,d),\\
        &gb[0] \notin S_{Loc}\} \\
        b(extra) = &\sum_{gb \in GB_{extra}} gb[1]
    \end{aligned}
\end{equation}

\begin{equation}\label{eq:map-b-3}
    \begin{aligned}
        & b(absb\_g) = 0.0 \\
        & b(absb\_ng) = 0.0
    \end{aligned}
\end{equation}
where $nodesAtHeight(B,d)$ returns the set of all nodes located at height $d$ in the global belief $B$ and $b(s)$ is the belief probability of state $s$ in the local state space $S_{Loc}$. Thus, for states that are not in $S_{Loc}$, the $mapBeliefToLocal$ procedure sums their probability in the $extra$ state, whereas for states that are in $S_{Loc}$, their probability is simply assigned to them from the GB. Once the initial belief distribution has been built, the value function (from which policy $\pi$ is derived) is modified (line 4) so that the resulting policy $\pi_E$ considers the dispersion of the probability of every state that is represented by the $extra$ state (the inner working of Algorithm \ref{alg:entropy-w}, which describes the $entropyWeight$ procedure, is detailed after the explanation of Algorithm \ref{alg:pomdp-pol-exec}).

\paragraph{}
Afterwards, the procedure enters in a loop in which it executes the last sampled action (lines 7-12), updates its local belief distribution (line 13), updates its policy based on the current entropy of the $extra$ state (line 14) and samples the next action. This loop will end when the action sampled by the policy is $terminate$ or $help$. Furthermore, since actions can either be concrete (those that take place at the bottom of the SST) or abstract, depending on which is the case, Algorithm \ref{alg:pomdp-pol-exec} will proceed differently. In the case of concrete actions (lines 8-9), the action is simply executed, and the perceived concrete observation is employed to update the GB $B$, using Algorithm \ref{alg:update-gb}. On the other hand, since abstract actions are POMDP policies, they are executed by recursively invoking Algorithm \ref{alg:pomdp-pol-exec}, in which the height passed as parameter is the current height plus one, \textit{i.e.}, $d+1$. Finally, the procedure returns an ordered pair containing the last sampled action and the current version of the the GB (line 17).

\begin{algorithm}
    \caption{Weighing of a value function}\label{alg:entropy-w}
    \begin{algorithmic}[1]
        \Procedure{entropyWeight}{$\pi$, $S_{Loc}$, $B$, $d$}
            \State $extra\_probs \gets [\:]$
            \State $i \gets 0$
            \ForAll{$gb$ at level $d$ in $B$}
                \If{$gb[0] \notin S_{Loc}$}
                    \State $append(extra\_probs, gb[1])$
                    \State $i \gets i + 1$
                \EndIf
            \EndFor
            \State $uni\_probs \gets uniformDistribution(i)$
            \State $E \gets entropy(extra\_probs)$
            \State $E_{max} \gets entropy(uni\_probs)$
            \State $V \gets getValueFunction(\pi)$
            \State $i \gets getStateIndex(S_{Loc},extra)$
            \ForAll{$\alpha$ in $V$}
                \State $\alpha[i] \gets \frac{\alpha[i]}{1 + |\alpha[i] * \frac{E}{E_{max}}|}$
            \EndFor
            \State $\pi_E \gets getPolicy(V)$
            \State \textbf{return} $\pi_E$
        \EndProcedure
    \end{algorithmic}
\end{algorithm}

\paragraph{}
With regards to Algorithm \ref{alg:entropy-w}, it receives as input parameters the unmodified policy ($\pi$), the state space of the policy ($S_{Loc}$), the current global belief ($B$) and the height in the SST at which the policy takes place ($d$). First, the probability of every state located at height $d$ that is not in $S_{Loc}$ is gathered to build the probability distribution $extra\_probs$ (lines 2-9), followed by the construction of a uniform distribution with the same amount of elements as $extra\_probs$ (line 10). Then, the Shannon entropy of both distributions is computed (lines 11-12) to weight the value of the $extra$ state in every $\alpha$-vector of the value function from which $\pi$ was computed. Finally, policy $\pi_E$ is computed from the value function after the entropy-based modification.

\paragraph{}
The value function is modified because, when policy $\pi$ was computed, the POMDP solving algorithm did not know that the $extra$ state actually represents a set of states. Thus, without the modification Algorithm \ref{alg:entropy-w} performs, policy $\pi$ would sample actions under the assumption that each probability in the belief distribution is concentrated in a single state of the environment, which may not be necessarily true for the $extra$ state. Hence, in Algorithm \ref{alg:entropy-w}, we use the entropy of the probability distribution of the states that $extra$ represents, to measure the degree in which such assumption is true for the $extra$ state (line 16). Where, $E$ is the current entropy value for the $extra$ state and $E_{max}$ is the maximum entropy value the $extra$ state can have (a uniform distribution). Therefore, when the probability of the $extra$ state is allocated in a single state, $E$ will equal zero and the value of the $extra$ state in the $\alpha$-vectors will remain the same. However, as the dispersion of the probability of the $extra$ state gets closer to a uniform distribution, the larger $E$ will be, and the more diluted the value of the $extra$ state will be in the $\alpha$-vectors.

\begin{algorithm}
    \caption{Execution of a hierarchical policy}\label{alg:hie-pol-exec}
    \begin{algorithmic}[1]
        \Procedure{ExecuteHiePolicy}{$\Pi^H$, $b_0$}
            \State $B \gets buildGlobalBelief(b_0)$
            \State $i \gets 0$
            \While{$i < length(\Pi^H)$}
                \State $[a,B] \gets ExecutePolicy(\Pi^H[i],B,i+1)$
                \If{$a = terminate$}
                    \State $i \gets i + 1$
                \Else
                    \If{$last\_a = help$}
                        \State $i \gets i - 1$
                    \EndIf
                \EndIf
            \EndWhile
        \EndProcedure
    \end{algorithmic}
\end{algorithm}

\paragraph{}
Regarding the execution of an HP, Algorithm \ref{alg:hie-pol-exec} takes as input parameters the HP to be executed ($\Pi^H$) and the initial belief state distribution ($b_0$) for the states located at the bottom of the SST (concrete states). First, a GB is built from $b_0$ by performing the steps of Algorithm \ref{alg:update-gb}, without updating the bottom level of a GB. In other words, the probability of every concrete state is pushed upwards to its ancestor nodes. Next, the procedure enters into a loop in which the first LP ($\Pi^H[0]$) takes places at the level below the root of the SST, while the last LP ($\Pi^H[length(\Pi^H) - 1]$) takes place at the bottom of the SST. The execution of the LPs is done by invoking Algorithm \ref{alg:pomdp-pol-exec}, and when its execution ends, the returned action will determine the next LP to be executed (Fig. \ref{fig:h-pol} shows an example of an HP).

\paragraph{}
That is, if the last action an LP performed was $terminate$, it is likely that the LP believed it had reached its goal state, therefore, Algorithm \ref{alg:hie-pol-exec} would proceed to execute the LP in the next level below in the SST. However, if its last action was $help$ it might be due to the LP believing the agent had reached the $extra$ state, meaning that the agent left the local state space of the LP. In this case, the LP from the level above is executed, since its local state space is broader and, hence, it is likely that the agent is within it.

\begin{figure*}[t!]
    \centering
    \includegraphics[width=\textwidth]{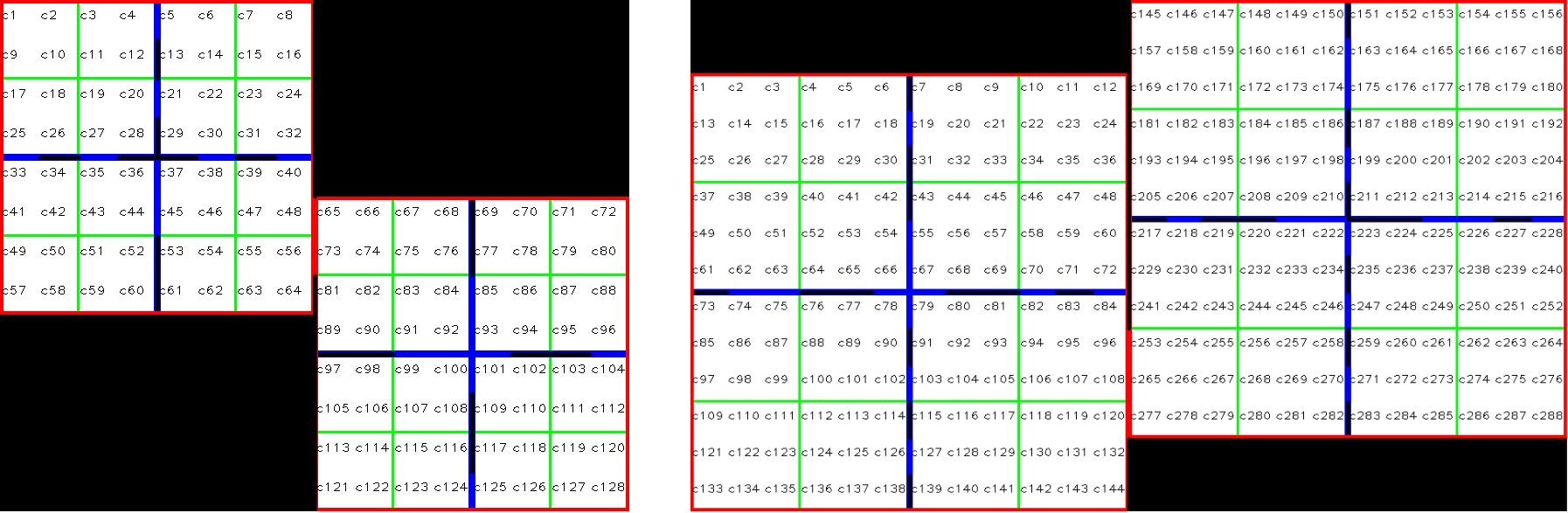}
    \caption{Navigation environments generated for the two first experimental configurations evaluated in the set of experiments 3 (see Fig. \ref{fig:exp3}). Sections, rooms and buildings are framed in green, blue and red, respectively. The black segments, which overlap with some of the blue and red lines, represent walls that block the way of the robot.}
    \label{fig:envs}
\end{figure*}

\begin{figure*}[t!]
    \centering
    \includegraphics[width=\textwidth]{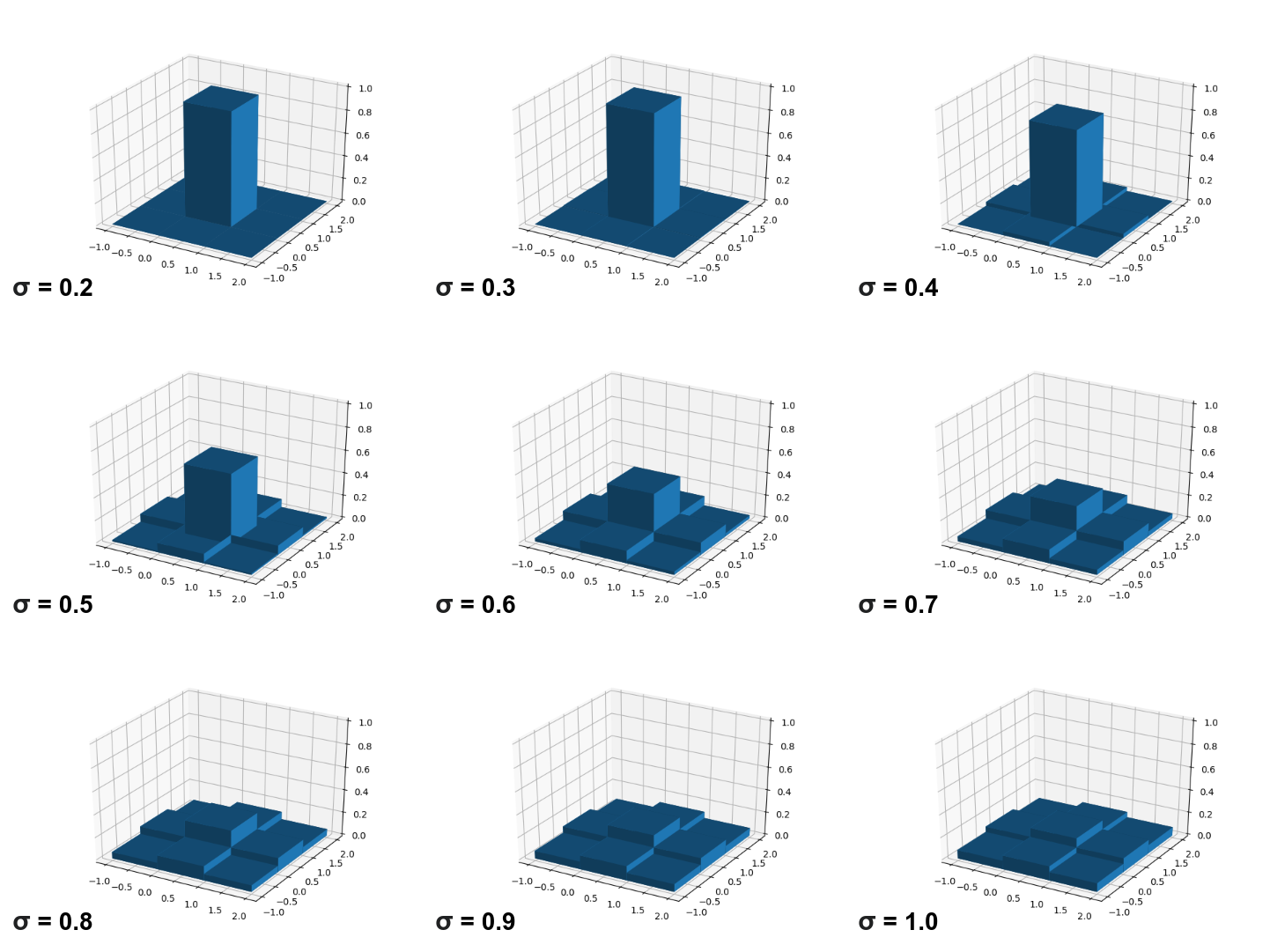}
    \caption{3D histograms of the discrete Gaussian $3 \times 3$ kernels (with different standard deviations) employed to model the observation distribution for the concrete actions in our experiments. The greater the probability for the bin in the center, the less likely it is for the agent to perceive misleading observations. On the other hand, the flatter the histogram is, the greater the noise is in the observation function.}
    \label{fig:o-kernels}
\end{figure*}

\section{Experiments} \label{sec:exp}
In order to evaluate the proposed architecture, a mobile robot navigation domain was employed. In this domain, an environment was modeled as a sequence of interconnected buildings, each one discretized as a grid of square uniform cells (Fig. \ref{fig:envs} shows two environments). The bottom POMDP, obtained from the description of the domain encoded in the knowledge base, has a state and an observation for each cell in the environment, while the set of actions is constituted by: \textit{up, down, left, right}. The transition distribution of each action assigns probabilities of 0.1 and 0.9 to staying in the current cell and transiting to the target cell, respectively (see Fig. \ref{fig:nav-t}). The observation distribution of each action is modeled as a $3 \times 3$ Gaussian kernel centered in the reached cell (see Fig. \ref{fig:nav-t}), whose standard deviation is specified differently for several experimental configurations (see Fig. \ref{fig:o-kernels}). For the hierarchical function, four levels are provided (from bottom to the top of the hierarchy): cells, sections, rooms and buildings.

\paragraph{}
For comparison purposes, the proposed architecture (HP) is compared against a standard POMDP (FP) and a two level hierarchical approach (TLP), that in an initialization phase computes POMDP policies designed to transit between buildings. When a task request is received, it computes a buildings path, executes a sequence of building-to-building policies to traverse the path and, once it believes it has reached the building that contains the goal cell, computes a POMDP policy whose set of states is made of all the cells that are in the goal building and proceeds to execute it. Thus, TLP requires to know the building at which the agent is at the beginning of the task. Also, in order to solve the POMDP policies for FP, TLP and HP, the point based value iteration algorithm \cite{pineau2003point} was used, whereas, 100 simulations were performed by HP to estimate the transition distribution of each abstract action.

\paragraph{}
Decomposing a task into a set of smaller ones usually makes easier to compute a solution for it, however, it also comes with a cost in terms of the quality of the solution. When one of the decomposed smaller tasks is being solved, since it is treated as an isolated problem, its solution is computed with less information (in comparison to solving the original problem), which may affect its performance in the long term. To quantify how easy it is for a method to compute a solution we use the time required to plan (seconds). Whereas, to measure the quality of the solution we employ Eqs. (\ref{eq:ev-sr}), (\ref{eq:ev-rc}) and (\ref{eq:ev-re}). Equation (\ref{eq:ev-sr}) quantifies the ratio of runs in which the goal cell was reached with respect to the total runs. Equation (\ref{eq:ev-rc}) computes how many concrete actions the agent took to arrive to the goal cell, and Eq. (\ref{eq:ev-re}) measures how far the agent is from the goal cell at the end of a run. $SP(\cdot,\cdot)$ computes the length (in horizontal and vertical steps) of the shortest path between two cells.
\begin{equation}\label{eq:ev-sr}
    success\:ratio = \frac{\#\:successful\:runs}{total\:runs}
\end{equation}

\begin{equation}\label{eq:ev-rc}
    path\:relative\:cost = \frac{\#\:concrete\:actions}{SP(initial\:cell,goal\:cell)}
\end{equation}

\begin{equation}\label{eq:ev-re}
    relative\:error = \frac{SP(final\:cell,goal\:cell)}{SP(initial\:cell,goal\:cell)}
\end{equation}

\paragraph{}
The evaluation process has been structured in three sets of experiments, which are summarized in Table \ref{tab:exp-summ}. Each experimental configuration is defined by the height and width of all sections, rooms and buildings, the standard deviation used in the Gaussian kernel that models the observation distributions of each action, the initial belief distribution the agent has at the beginning of a task and the amount of buildings. All sections, rooms and building are square, and their dimensions are expressed in terms of the objects in the level below in the hierarchy (\textit{e.g}, sections in terms of cells and rooms in terms of sections, see Fig. \ref{fig:envs} for some environment examples). A total amount of twenty three experimental configurations were evaluated (see Table  \ref{tab:exp-config}), all of them used an environment made of two buildings.

\begin{table}[]
\centering
\caption{Summary of the sets of experiments.}
\label{tab:exp-summ}
\begin{tabular}{ll}
\hline
\begin{tabular}[c]{@{}l@{}}Set of\\ experiments\end{tabular} & Description \\ \hline
1 & \begin{tabular}[c]{@{}l@{}}The agent knows with certainty its initial\\ state, the environment is made of 128\\ cells and the standard deviation was va-\\ ried, results are shown in Fig. \ref{fig:exp1}.\end{tabular} \\
2 & \begin{tabular}[c]{@{}l@{}}The agent starts with a uniform distri-\\ bution as initial belief state, the envi-\\ ronment is made of 128 cells and the\\ standard deviation was varied, results\\ are shown in Fig. \ref{fig:exp2}.\end{tabular} \\
3 & \begin{tabular}[c]{@{}l@{}}The agent knows with certainty its initial\\ state, the standard deviation is fixed to \\ 0.2 while the dimensions of the sections,\\ rooms and buildings are were varied, re-\\ sults are shown in Fig. \ref{fig:exp3}.\end{tabular} \\ \hline
\end{tabular}
\end{table}

\begin{table*}[t!]
\centering
\caption{Specification of the experimental configurations evaluated in the sets of experiments 1, 2 and 3. Each row corresponds to a configuration, while columns (from left to right) show: the set of experiments to which  the configuration belongs to, the height and width of every section, room and building, the standard deviation used to compute the Gaussian kernel that models the observations distributions, and the belief distribution the agents starts with in every task (where \textit{uniform} refers to a uniform distribution and $b(s_0) = 1.0$ means that the agent knows its initial state with certainty). The bold values correspond to the non-fixed parameters in each set of experiments.}
\label{tab:exp-config}
\begin{tabular}{llllll}
\hline
\begin{tabular}[c]{@{}l@{}}Set of\\ experiments\end{tabular} & \begin{tabular}[c]{@{}l@{}}Section\\ dimensions\\ (cells)\end{tabular} & \begin{tabular}[c]{@{}l@{}}Room\\ dimensions\\ (sections)\end{tabular} & \begin{tabular}[c]{@{}l@{}}Building\\ dimensions\\ (rooms)\end{tabular} & \begin{tabular}[c]{@{}l@{}}Standard deviation\\ in Gaussian kernel\end{tabular} & \begin{tabular}[c]{@{}l@{}}Initial\\ belief\\ distribution\end{tabular} \\ \hline
1 & 2 & 2 & 2 & \textbf{0.2} & $b(s_0) = 1.0$ \\
1 & 2 & 2 & 2 & \textbf{0.3} & $b(s_0) = 1.0$ \\
1 & 2 & 2 & 2 & \textbf{0.4} & $b(s_0) = 1.0$ \\
1 & 2 & 2 & 2 & \textbf{0.5} & $b(s_0) = 1.0$ \\
1 & 2 & 2 & 2 & \textbf{0.6} & $b(s_0) = 1.0$ \\
1 & 2 & 2 & 2 & \textbf{0.7} & $b(s_0) = 1.0$ \\
1 & 2 & 2 & 2 & \textbf{0.8} & $b(s_0) = 1.0$ \\
1 & 2 & 2 & 2 & \textbf{0.9} & $b(s_0) = 1.0$ \\
1 & 2 & 2 & 2 & \textbf{1.0} & $b(s_0) = 1.0$ \\
2 & 2 & 2 & 2 & \textbf{0.2} & Uniform \\
2 & 2 & 2 & 2 & \textbf{0.3} & Uniform \\
2 & 2 & 2 & 2 & \textbf{0.4} & Uniform \\
2 & 2 & 2 & 2 & \textbf{0.5} & Uniform \\
2 & 2 & 2 & 2 & \textbf{0.6} & Uniform \\
2 & 2 & 2 & 2 & \textbf{0.7} & Uniform \\
2 & 2 & 2 & 2 & \textbf{0.8} & Uniform \\
2 & 2 & 2 & 2 & \textbf{0.9} & Uniform \\
2 & 2 & 2 & 2 & \textbf{1.0} & Uniform \\
3 & \textbf{2} & \textbf{2} & \textbf{2} & 0.2 & $b(s_0) = 1.0$ \\
3 & \textbf{3} & \textbf{2} & \textbf{2} & 0.2 & $b(s_0) = 1.0$ \\
3 & \textbf{3} & \textbf{3} & \textbf{2} & 0.2 & $b(s_0) = 1.0$ \\
3 & \textbf{3} & \textbf{3} & \textbf{3} & 0.2 & $b(s_0) = 1.0$ \\ \hline
\end{tabular}
\end{table*}

\begin{figure*}[h!]
    \centering
    \includegraphics[width=0.7\textwidth]{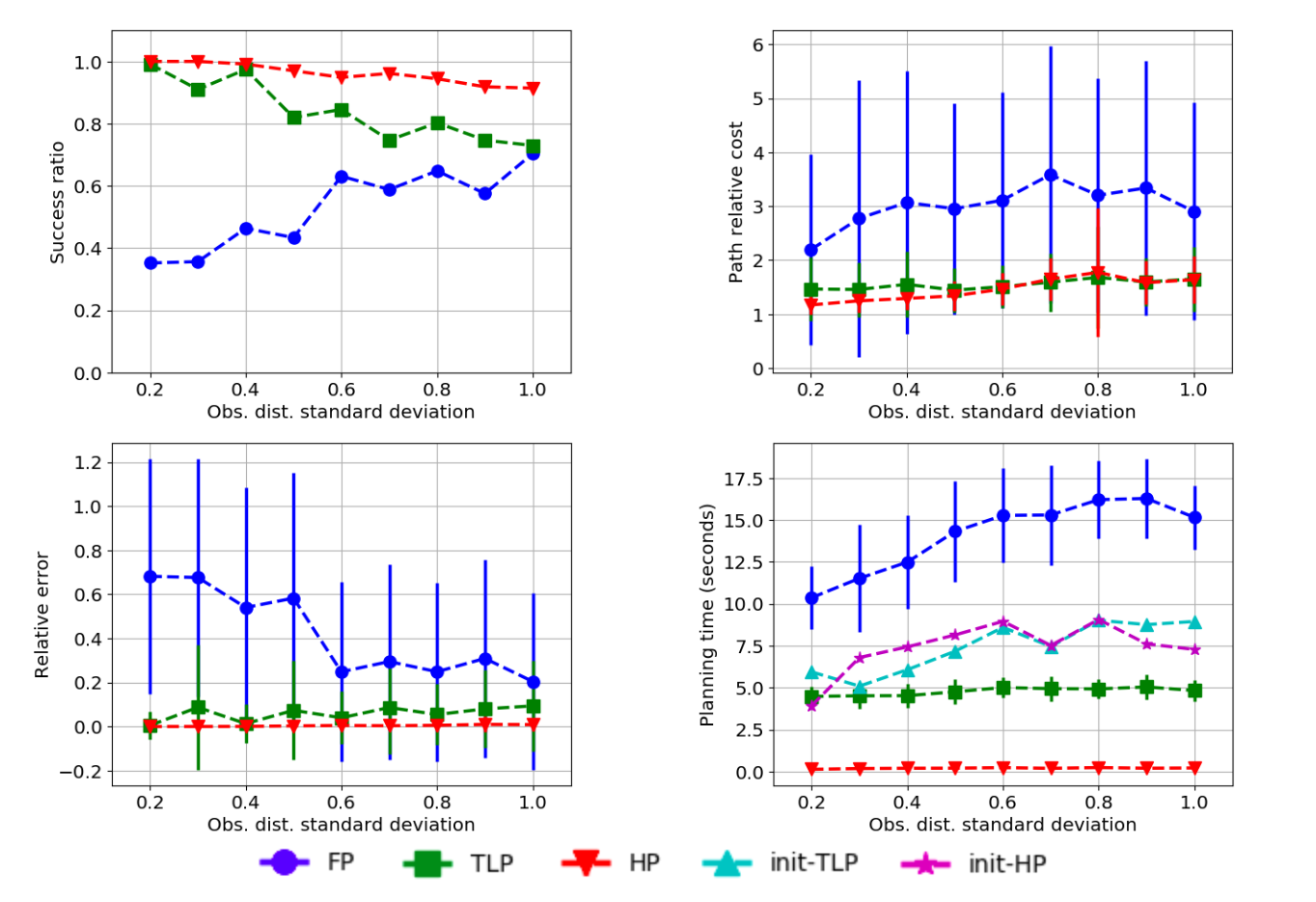}
    \caption{Results for the set of experiments 1. The vertical axes show the average and standard deviation (except for the success ratio) obtained in each evaluation metric (from left to right and top to bottom): success ratio (Eq. (\ref{eq:ev-sr})), path relative cost (Eq. (\ref{eq:ev-rc})), relative error (Eq. (\ref{eq:ev-re})) and planning time, whereas the horizontal axes correspond to the standard deviation employed to compute the Gaussian kernel that models the observation distributions. The init-TLP and init-HP labels represent the planning time required for the initialization stages of the TLP baseline method and the proposed architecture (HP), respectively.}
    \label{fig:exp1}
\end{figure*}

\begin{figure*}[h!]
    \centering
    \includegraphics[width=0.7\textwidth]{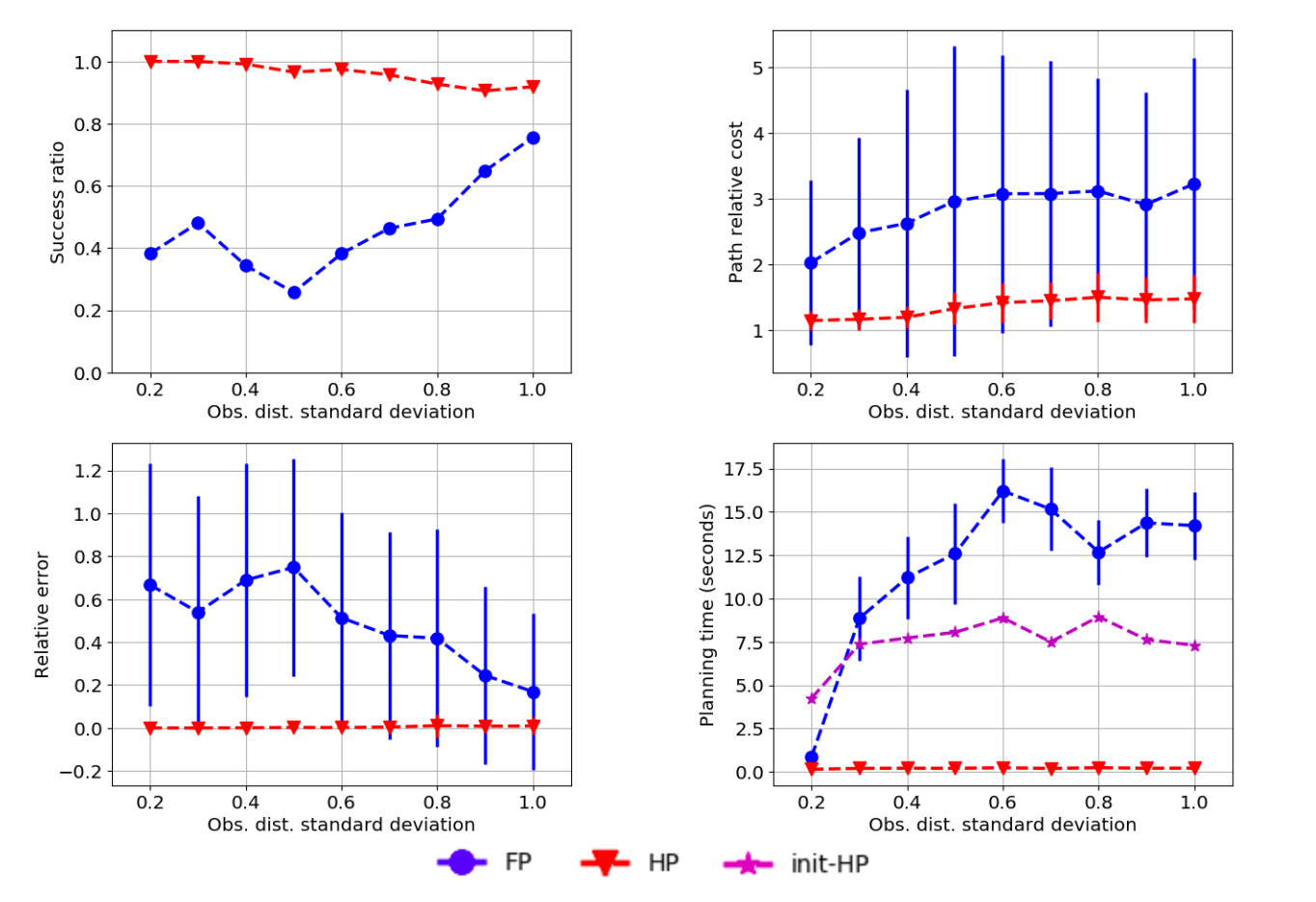}
    \caption{Results for the set of experiments 2. The vertical axes show the average and standard deviation (except for the success ratio) obtained in each evaluation metric (from left to right and top to bottom): success ratio (Eq. (\ref{eq:ev-sr})), path relative cost (Eq. (\ref{eq:ev-rc})), relative error (Eq. (\ref{eq:ev-re})) and planning time, whereas the horizontal axes correspond to the standard deviation employed to compute the Gaussian kernel that models the observation distributions. The init-HP label represents the planning time required for the initialization stage of the proposed architecture (HP).}
    \label{fig:exp2}
\end{figure*}

\begin{figure*}[h!]
    \centering
    \includegraphics[width=0.7\textwidth]{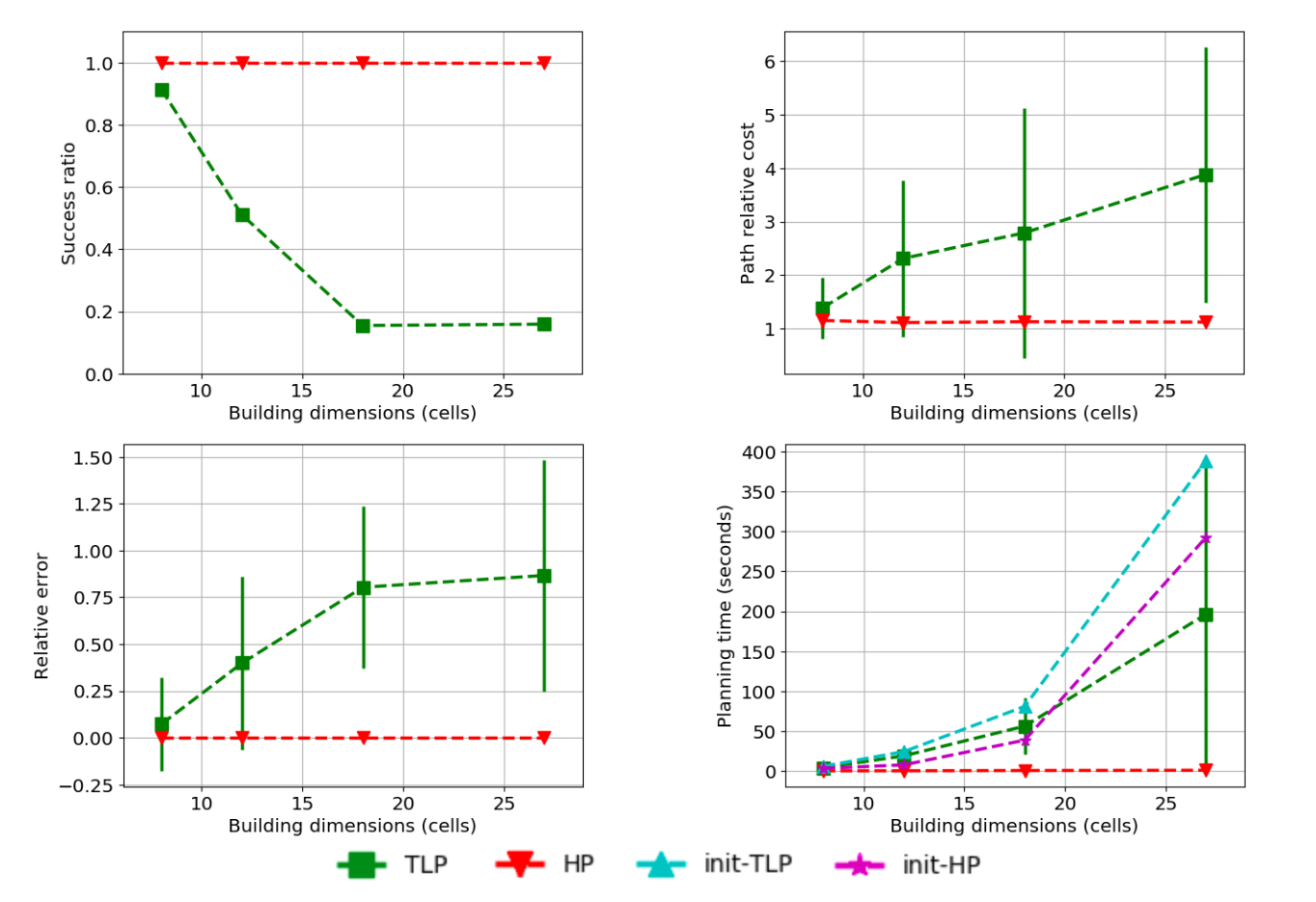}
    \caption{Results for the set of experiments 3. The vertical axes show the average and standard deviation (except for the success ratio) obtained in each evaluation metric (from left to right and top to bottom): success ratio (Eq. (\ref{eq:ev-sr})), path relative cost (Eq. (\ref{eq:ev-rc})), relative error (Eq. (\ref{eq:ev-re})) and planning time, whereas the horizontal axes correspond to the width and height (in cells) of each one of buildings that constitute the environment. The init-TLP and init-HP labels represent the planning time required for the initialization stages of the TLP baseline method and the proposed architecture (HP), respectively.}
    \label{fig:exp3}
\end{figure*}

\paragraph{}
For every experimental configuration, 233 runs were performed. In each run, a pair of initial and goal cells was randomly sampled, each cell from one of the buildings that constitute the environment. Figure \ref{fig:envs} shows some examples of the environments generated for the experiments, whereas, Figs. \ref{fig:exp1}, \ref{fig:exp2} and \ref{fig:exp3} show the results for the sets of experiments 1, 2 and 3, respectively. Furthermore, it is worth noting that most of the general knowledge (except the observation probabilities) was reused throughout all the experiments, and only the specific knowledge had to be specified for each environment (\textit{i.e.}, for each experimental configuration). The general-specific partition of the knowledge facilitated the process of designing the evaluation setting, as the robot was described independently from the test environments.

\paragraph{}
As the plots from Fig. \ref{fig:exp1} show, both TLP and HP were able to consistently reach the goal cell in most of the runs, while FP was not capable of scoring a success ratio greater than $0.8$ in any experimental configuration. Even though an observation function with greater uncertainty did not seem to affect the robustness of TLP and HP, it did increase the average path relative cost in the last four configurations. This may be because the agent perceived, in a greater rate, observations that misinformed it about its true location, which led it to take actions that derailed it from the shortest path to its goal. Furthermore, despite TLP and HP showed a similar performance in terms of effectiveness, HP is considerably less time consuming.

\paragraph{}
Since in the second set of experiments the initial belief was a uniform distribution, TLP was not evaluated, because it requires to know the initial state to compute the path of buildings. Thus, for this set of experiments we observed a very similar behavior in FP and HP (with respect to the previous set). It is likely that the high degree of uncertainty, at the beginning of a task, did not affect the performance of neither methods because the observation distribution of every concrete action is defined over a local neighborhood. Hence, the agent required a short sequence of observations to compute a good estimate of its location.

\paragraph{}
For the third set of experiments, we did not evaluate FP since it had already shown to not perform well in small environments with low uncertainty. With regards to TLP and HP, for the first configuration they performed quite similar. However, as the size of the environments increased in the last three configurations, the success ratio of TLP significantly dropped, whereas, HP maintained its effectiveness. Furthermore, although the initial planning time required by HP (the time it takes to build  the hierarchy of actions) grows non-linearly, the time required to build a hierarchical policy shows a tendency to remain constant. Hence, the larger the period of operation is, the more the robot will gain from the time invested in the initial planning, and from the hierarchical knowledge available.

\subsection{Discussion}
Since the main motivation of our work is to endow a robot with the capacity to solve tasks, efficiently and effectively, we selected a single-domain scenario (navigation with a mobile robot), as it enabled us to study the proposed methodology and the effect certain variables (like the size of the problem and the dispersion in the observation distribution) had in the behavior of the system, without the interference that the interplay among a set state variables might have inserted. As the red trace (from the planning-time plots) shows in Figs. \ref{fig:exp1}, \ref{fig:exp2} and \ref{fig:exp3}, our architecture requires a considerably less time to generate the plan for a task request than the baseline methods. As it should, given the advantage it has by knowing the underlying hierarchical structure of the environment.

\paragraph{}
However, we were less certain about the effect the uncertainty in the observation distributions would have in the system, as the abstract actions operate over small portions of the state space. Our initial hypothesis was that, as the uncertainty increased, the system would find itself in the $extra$ state more frequently, and that it would get stuck in its execution, passing the control from one abstract action to the other, without really doing anything. Nonetheless, passing the control to the parent policy (whether it is an LP or an abstract action) seems to help the system reevaluate and invoke a better action, given the current belief state.

\paragraph{}
On the other hand, none of the environments employed in the experiments had dead ends within local state spaces (see Fig. \ref{fig:envs}), \textit{i.e.}, states that could not be reached with any sequence of actions from other states within the same local state space. This sort of scenarios are still a challenge for hierarchical approaches, as local POMDPs ignore information from other regions of the state space, and might lead to sub-optimal behaviors. Currently, our architecture does not have a strategy to compensate for the lack of information in the local POMDPs. However, in a similar way the entropy-based weight brings information about the outer state space in a local POMDP, abstract states could share, with their neighborhood, statistics that summarize relevant information about them and that improve the decision making of local POMDPs.

\paragraph{}
Moreover, another challenge yet to be overcome is to design a reward function that represents the actual costs of actions. In general, designing reward functions that actually express the behavior we want the agent to show is a hard problem in reinforcement learning. Alternatives have been explored, such as inverse reinforcement learning \cite{ng2000algorithms} that attempts to learn the reward function from a set of demonstrations from an expert on the task at hand. Although there is still no clear answer to this problem, there are some things we can do so that, during the construction of the hierarchy of POMDPs, the architecture takes into account the reward function from lower levels in the SST. For instance, we could count the amount of actions taken during the simulation of policies that belong to abstract actions, as a way to measure the efficiency (in steps) of an abstract action. Hence, the efficiency value could be transformed into a reward value that describes how costly an abstract action is.

\paragraph{}
Although there are still aspects from our architecture that can significantly be improved, by building of a hierarchy of actions (POMDPs) based on an abstraction of the state space, the architecture has the capacity to exploit hierarchical information in (virtually) any sort of task planning scenario. Being task diversity a common characteristic in service robotics \cite{ingrand2017deliberation}, adaptability becomes a critical feature in planning systems. Contrary to works that start from a hierarchy of abstract actions provided by a designer (\textit{e.g.}, \cite{pineau2001hierarchical} and \cite{pineau2002integrated}), our architecture builds abstract actions that fit into the particularities of the environment that the robot will encounter. Hence, no assumptions are made about which actions are necessary for certain abstract action, and configurations that one could regard unexpected (or exceptions to the rule) are considered in the construction of the hierarchy of actions.

\paragraph{}
Additionally, as in any decision-making system, it is important to have a way for us (users) to monitor the reasoning that is behind the actions we observe. Furthermore, as larger problems with many state variables are encountered, the behavior of standard frameworks like MDP and POMDP become almost impossible to interpret. In contrast, the proposed architecture facilitates the readability of why the agent executes certain action. That is, in the execution of a hierarchical policy, the control is passed between LPs up and down. The direction in which the control is passed, can be interpreted as follows: i) the control is passed upwards to gain a broader panorama of the current situation, and ii) the control is passed downwards when the agent gets closer to its goal, and a more focused vision is required to get the work done. That is, passing the control up and down can be seen as the way the agent constantly adjusts the region of the problem it considers it should focus on.


\section{Conclusions and future work} \label{sec:conc-fw}
In this paper, we proposed a task planning architecture that automatically builds a hierarchy of POMDPs. This was achieved by introducing a new recursive definition for modeling POMDPs, based on a hierarchical description of the state space. For a given encoding of the robot and the environment (provided by a designer), our architecture builds a hierarchy of POMDPs that is employed to generate and execute plans for specific task requests. The execution of such plans is driven by the entropy of the belief state of the agent, enabling it to focus in small regions of the state space when it is sure about its current state, as well to reconsider (at a more abstract level) what action should be taken if otherwise. The main advantages of our architecture are: i) its recursive definition enables the construction of hierarchies of POMDPs of any depth (depends on the description provided by the designer), ii) once the KB has been specified, human intervention is not required to build the hierarchy of POMDPs, but rather to issue task requests, iii) by describing separately the skills of a robot, the features all environments have, as well as the particularities they might not share, the architecture promotes reusing knowledge and designing robots in a modular-incremental fashion.

\paragraph{}
In fact, despite that we evaluated the architecture in a single-domain environment (navigation), it is worth noting that as the amount of basic modules and environments increase, the greater the impact the knowledge base will have on the robot design process. That is, for a robot whose hardware is still under development, new skills could be encapsulated in basic modules and seamlessly incorporated in an already functional architecture. On the other hand, in the case of a float of robots that is about to be deployed in various households, the general knowledge could be reused and only the specific knowledge would require to be encoded in each unit. Thus, in addition to considering the scale-up factor with regards to the size of the planning problems, our architecture provides a scheme that scales up in terms of human effort.

\paragraph{}
As for its performance, experimental results show the robustness of the architecture in solving tasks across a spectrum of noise levels in the observation distribution, as well in scenarios where the initial state is not known. Furthermore, the results from the set of experiments 3 (see Fig. \ref{fig:exp3}) exhibit how dramatically the planning time can be reduced when a system is capable of exploiting all the hierarchical information available. On the other hand, there are several directions in which the proposed architecture could be improved, for instance, by integrating commonsense reasoning capabilities, similar to the work presented in \cite{zhang2020icorpp}. In this way, the architecture could employ commonsense knowledge to model information that is hard to encode in a POMDP (\textit{e.g.}, preferences of the user, places where objects usually are, etc.), use this information to compute plans as sequences of high-level actions, and perform each action as an HP. In addition, we would like to evaluate our architecture in a more intricate scenario than the navigation one, and observe how well the system performs in an environment described by several basic modules.

\begin{acknowledgements}
Elizabeth Santiago deeply thanks to the postdoctoral scholarship and Sergio A. Serrano thanks the scholarship No. 489044, both granted by CONACYT, and also to the Robotics Laboratory of the National Institute of Astrophysics, Optical and Electronic which were important for the realization of this work.
\end{acknowledgements}
%
\section*{Declarations}
\textbf{Funding}: Not applicable.\\
\textbf{Conflicts of interest}: The authors declare that they have no conflict of interest.\\
\textbf{Availability of data and material}: Not applicable.\\
\textbf{Code availability}: Not applicable.\\
\textbf{Authors' contributions}: All authors contributed to the study conception, design and implementation. All authors read and approved the final manuscript.
%

\bibliographystyle{abbrv}
\bibliography{references}   

\end{document}